\documentclass[11pt,a4paper]{article}
\usepackage[hyperref]{AACL-IJCNLP2020}
\usepackage{times}
\usepackage{latexsym}
\usepackage{graphicx}
\usepackage{adjustbox}
\usepackage{xspace}
\usepackage{booktabs}
\usepackage{amssymb}
\usepackage{pifont}
\usepackage{multirow}
\newcommand{\cmark}{\ding{51}}%
\newcommand{\xmark}{\ding{55}}%

\usepackage[hyphenbreaks]{breakurl}  

\newcommand{\breakableurl}[1]{\texttt{#1}}
\usepackage{microtype}

\aclfinalcopy 
\def\aclpaperid{275} 

\title{Multimodal Pretraining for Dense Video Captioning}

\author{Gabriel Huang\textsuperscript{1}\thanks{\, This work was done while Gabriel Huang was an intern at Google Research.},
\, Bo Pang\textsuperscript{2},
\,Zhenhai Zhu\textsuperscript{2},
\,Clara Rivera\textsuperscript{2},
\,Radu Soricut\textsuperscript{2}\\
\textsuperscript{1}Mila \& University of Montreal\\
\textsuperscript{2}Google Research\\
\small
\texttt{gabriel.huang@umontreal.ca}\\
\small
\texttt{\{bopang,zhenhai,rivera,rsoricut\}@google.com }
}

\date{}

\begin{document}
\maketitle

\newcommand{\ldvm}{ViTT\xspace}
\newcommand{\ldvmlong}{Video Timeline Tags\xspace}
\newcommand{\ldvmmerged}{\ldvm-All\xspace}
\newcommand{\ldvmcooking}{\ldvm-Cooking\xspace}
\newcommand{\youcook}{YouCook2\xspace}
\newcommand{\ytdata}{YouTube-8M\xspace}
\newcommand{\ytdatasub}{YT8M-cook\xspace}
\newcommand{\youtube}{\ytdatasub\xspace}
\newcommand{\howtodata}{HowTo100M\xspace}
\newcommand{\howto}{HowTo100M\xspace}
\newcommand{\recipes}{Recipe1M\xspace}
\newcommand{\wikihow}{WikiHow\xspace}
\newcommand{\starburst}{Compact 2D\xspace}
\newcommand{\fnet}{FDense\xspace}
\newcommand{\unimt}{UniD\xspace}
\newcommand{\bimt}{BiD\xspace}
\newcommand{\bimtalt}{BiDalt\xspace}
\newcommand{\rouge}{{\sc{Rouge}}-L\xspace}
\newcommand{\bleu}{{\sc{Bleu}}\xspace}
\newcommand{\meteor}{{\sc{Meteor}}\xspace}
\newcommand{\cider}{{\sc{CIDEr}}\xspace}
\newcommand{\ASRtoASR}{\textsc{asr}$\to$\textsc{asr}\xspace}
\newcommand{\CAPtoCAP}{\textsc{cap}$\to$\textsc{cap}\xspace}
\newcommand{\ex}[4]{\adjustimage{width=0.19\textwidth,valign=M,frame}{#1}%
&%
\textit{#2}%
&%
\textit{#3}%
&%
\textit{#4}%
}
\newcommand{\good}{\textcolor{green!70!black}{(\textbf{good})}}
\newcommand{\ok}{\textcolor{orange}{(\textbf{ok})}}
\newcommand{\bad}{\textcolor{red}{(\textbf{bad})}}

\begin{abstract}
Learning specific hands-on skills such as cooking, car maintenance, and home repairs increasingly happens via instructional videos.
The user experience with such videos is known to be improved by meta-information such as time-stamped annotations for the main steps involved.
Generating such annotations automatically is challenging, and we describe here two relevant contributions.
First, we construct and release a new dense video captioning dataset, \textbf{Vi}deo \textbf{T}imeline \textbf{T}ags (\ldvm), featuring a variety of instructional videos together with time-stamped annotations.
Second, we explore several multimodal sequence-to-sequence pretraining strategies that leverage large unsupervised datasets of videos and caption-like texts.
We pretrain and subsequently finetune dense video captioning models using both YouCook2 and ViTT.
We show that such models generalize well and are robust over a wide variety of instructional videos.
\end{abstract}

\section{Introduction}
YouTube recently reported that a billion hours of videos were being watched on the platform every day~\citep{youtubeblog}.
In addition, the amount of time people spent watching online videos was estimated to grow at an average rate of 32\% a year between 2013 and 2018, with an average person forecasted to watch 100 minutes of online videos per day in 2021~\citep{zenithmedia}.

An important reason for this fast-growing video consumption is information-seeking.
For instance, people turn to YouTube ``hungry for how-to and learning content'' \cite{oneilhart2018}.
Indeed, compared to traditional content format such as text, video carries richer information to satisfy such needs.
But as a content media, videos are also inherently more difficult to skim through, making it harder to quickly target the relevant part(s) of a video.
Recognizing this difficulty, search engines started showing links to ``key moments'' within videos in search results, based on timestamps and short descriptions provided by the content creators themselves.\footnote{\breakableurl{https://www.blog.google/products/\\search/key-moments-video-search/}}
This enables users to get a quick sense of what the video covers, and also to jump to a particular time in the video if so desired.
This effort echoes prior work in the literature showing how users of instructional videos can benefit from human-curated meta-data, such as a timeline pointing to the successive steps of a tutorial \cite{kim2014crowdsourcing,margulieux2012subgoal,weir2015learnersourcing}.
Producing such meta-data in an automatic way would greatly scale up the efforts of providing easier information access to videos.
This task is closely related to the dense video captioning task considered in prior work~\cite{zhou2018towards,Zhou2018EndtoEndDV,krishna2017densecaptioning}, where an instructional video is first segmented into its main steps, followed by segment-level caption generation. 

\begin{figure}
\centering
\includegraphics[width=\linewidth]{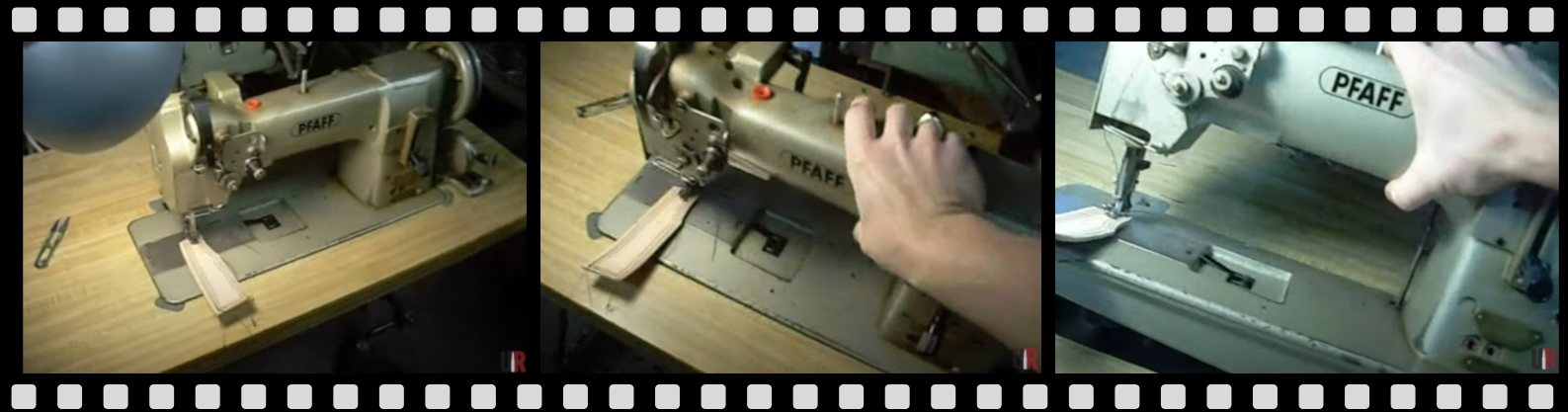}
{\small
\begin{tabular}{rl}
\textbf{Groundtruth} & \textit{Varying stiching speeds}\\    
\textbf{\O-Pretraining} & \textit{Showing other parts}   \\ 
\textbf{MASS-Pretraining} & \textit{Explaining how to do a stitch}   
\end{tabular}
}

\caption{Dense video captioning using \ldvm--trained models. For the given video scene, we show the \ldvm annotation (Groundtruth) and model outputs (no pretraining and MASS-based pretraining). }
\label{fig:my_label}
\end{figure}

To date, the \youcook data set~\cite{zhou2018towards} is the largest annotated data set for dense video captioning.
It contains annotations for 2,000 cooking videos covering 89 recipes, with per-recipe training / validation split.
Restricting to a small number of recipes is helpful for early exploratory work, but such restrictions impose barriers to model generalization and adoption that are hard to overcome.
We directly address this problem by constructing a larger and broader-coverage annotated dataset that covers a wide range of instructional topics (cooking, repairs, maintenance, etc.)
We make the results of our annotation efforts publicly available as \underline{\textbf{Vi}}deo \underline{\textbf{T}}imeline \underline{\textbf{T}}ags ({\bf \ldvm})\footnote{Available at \url{https://github.com/google-research-datasets/Video-Timeline-Tags-ViTT}}, consisting of around 8,000 videos annotated with timelines (on average 7.1 segments per video, each segment with a short free-text description).


Using \youcook and the new \ldvm dataset as benchmarks for testing model performance and generalization, we further focus on the sub-problem of video-segment--level caption generation, assuming segment boundaries are given ~\cite{hessel2019case,sun2019videobert,luo2020univilm}.
Motivated by the high cost of collecting human annotations, we investigate pretraining a video segment captioning model using unsupervised signals -- ASR (Automatic Speech Recognition) tokens and visual features from instructional videos, and {\em unpaired} instruction steps extracted from independent sources: Recipe1M~\cite{marin2019recipe1m+} and WikiHow~\cite{koupaee2018wikihow}.
In contrast to prior work that focused on BERT-style pretraining of encoder networks~\citep{sun2019videobert, sun2019contrastive},
our approach entails jointly pretraining both multimodal encoder and text-based decoder models via MASS-style pretraining~\cite{song2019mass}.
Our experiments show that pretraining with either text-only or multi-modal data provides significant gains over no pretraining, on both the established \youcook benchmark and the new \ldvm benchmark.
The results we obtain establish state-of-the-art performance on \youcook, and present strong performance numbers on the \ldvm benchmark.
These findings help us conclude that the resulting models generalize well and are quite robust over a wide variety of instructional videos.





\section{Related Work}
\label{sec:relwork}


\paragraph{Text-only Pretraining.} 
Language pretraining models based on the Transformer neural network architecture~\cite{Vaswani2017AttentionIA} such as  BERT~\citep{devlin2018bert}, GPT~\cite{GPT2018}, RoBERTa~\cite{Roberta},
MASS~\cite{song2019mass}
and ALBERT~\cite{Lan2020ALBERTAL} have achieved state-of-the-art results on many NLP tasks.
MASS~\citep{song2019mass} has been recently proposed as a joint encoder-decoder pretraining strategy. For sequence-to-sequence tasks, this strategy is shown to outperform approaches that separately pretrain the encoder (using a BERT-style objective) and the decoder (using a language modeling objective). 
UniLM~\cite{UniLM}, BART~\cite{Lewis2019BARTDS}, and T5~\cite{Raffel2019ExploringTL} propose unified pretraining approaches for both understanding and generation tasks.

\paragraph{Multimodal Pretraining.} 
VideoBERT~\cite{sun2019videobert}, CBT~\cite{sun2019contrastive} and ActBERT~\cite{actbert2020cvpr} use a BERT-style objective to train both video and ASR text encoders. \citet{Alayrac2016cvpr} and \citet{Miech2020cvpr} use margin-based loss functions to learn joint representations for video and ASR, and evaluate them on downstream tasks such as video captioning, action segmentation and anticipation, and action localization. 
An independent and concurrent work (UniViLM) by~\citet{luo2020univilm} is closely related to ours in that we share some similar pretraining objectives, some of the pretraining setup -- HowTo100M~\cite{Alayrac2016cvpr}, and the down-stream video captioning benchmark using \youcook~\cite{zhou2018towards}. 
The main difference is that they use BERT-style pretraining for encoder and language-modeling style pretraining for decoder, whereas we use MASS-style pre-training to pretrain encoder and decoder jointly.

Other approaches such as ViLBERT~\citep{lu2019vilbert}, LXMERT~\citep{tan2019lxmert}, Unicoder-VL~\citep{li2019unicoder}, VL-BERT~\citep{su2019vl}, and UNITER~\citep{chen2019uniter} focus on pretraining joint representations for text and image, evaluating them on downstream tasks such as visual question answering, image-text retrieval and referring expressions.

\paragraph{Dense Video Captioning.} 
In this paper, we focus on generating captions at the segment-level, which is a sub-task of the so-called dense video captioning task~\cite{krishna2017densecaptioning}, where fine-grained captions are generated for video segments, conditioned on an input video with pre-defined event segments. This is different from the video captioning models that generate a single summary for the entire video~\cite{Wang2019VaTeXAL}.

\citet{hessel2019case} make use of ASR and video for segment-level captioning on \youcook and show that most of the performance comes from ASR. \citet{shi2019dense,luo2020univilm} train their dense video captioning models on both video frames and ASR text and demonstrate the benefits of adding ASR as an input to the model. There are also a number of video captioning approaches that do not use ASR directly~\cite{Zhou2018EndtoEndDV, pan2020cvpr, zheng2020cvpr, zhang2020cvpr, Lei2020MARTMR}.

\paragraph{Instructional video captioning data sets.} 
In addition to \youcook~\cite{zhou2018towards}, there are two other smaller data sets in the instructional video captioning category. Epic Kitchen~\cite{DamenECCV2018} features 55 hours of video consisting of 11.5M frames, which were densely labeled for a total of 39.6K action segments and 454.3K object bounding boxes. How2~\cite{how2-dataset} consists of instructional videos with video-level (as opposed to segment-level) descriptions, authored by the video creators themselves. 

\section{Data}
\label{sec:data}

We present the datasets used for pretraining, finetuning, and evaluation in Table~\ref{table:datasets}.
We also describe in detail the newly introduced dense video captioning dataset, \underline{\textbf{Vi}}deo \underline{\textbf{T}}imeline \underline{\textbf{T}}ags (\ldvm).

\begin{table}
\centering
\begin{adjustbox}{width=\linewidth}
\begin{tabular}{llr}
\toprule
\textbf{Name} & \textbf{Type} & \textbf{\# segments}\\
\midrule
\multicolumn{3}{l}{\textit{Pretraining datasets}} \\
\ytdatasub & ASR+video & 186 K\\
\howto & ASR+video & 8.0 M\\
\recipes & CAP-style & 10.8 M\\
\wikihow & CAP-style & 1.3 M\\
\midrule
\multicolumn{3}{l}{\textit{Finetuning datasets}} \\
\youcook & ASR+video+CAP & 11.5 K\\  
\ldvmmerged & ASR+video+CAP & 88.5 K\\   
\bottomrule
\end{tabular}
\end{adjustbox}
\caption{Datasets used in this work, along with size of the data measured by the total number of segments. \label{table:datasets}}
\end{table}

\subsection{Dense Video-Captioning Datasets}
\label{sec:data:labeled}

Our goal is to generate captions (CAP) for video segments. We consider two datasets with segment-level captions for fine-tuning and evaluating ASR+Video$\to$CAP models. 

\paragraph{\youcook.} Up to this point, \youcook~\citep{zhou2018towards} has been the largest human-annotated dense-captioning dataset of instructional videos publicly available.  It originally contained 2,000 cooking videos from YouTube.
Starting from 110 recipe types (e.g., ``shrimp tempura''), 25 unique videos per recipe type were collected; the recipe types that did not gather enough videos were dropped, resulting in a total of 89 recipe types in the final dataset. 
In addition, \newcite{youcook2dataset} ``randomly split the videos belonging to each recipe into 67\%:23\%:10\% as training, validation and test sets\footnote{Note that no annotations are provided for the test split; we conducted our own training/dev/test split over available videos.},'' which effectively guarantees that videos in the validation and test sets are never about unseen recipes.  
Annotators were then asked to construct recipe steps for each video --- that is, identify the start and end times for each step, and provide a recipe-like description of each step.
Overall, they reported an average of 7.7 segments per video, and 8.8 words per description.  
After removing videos that had been deleted by users, we obtained a total of 11,549 segments.


\paragraph{\ldvm.} One limitation of the \youcook dataset is the artificially imposed (almost) uniform distribution of videos over 89 recipes.  While this may help making the task more tractable, 
it is difficult to judge whether performance on its validation / test sets can be generalized to unseen topics.

The design of our \ldvm dataset annotation process is aimed at fixing some of these drawbacks.
We started by collecting a large dataset of videos containing a broader variety of topics to better reflect topic distribution in the wild.
Specifically, we randomly sampled instructional videos from the \ytdata dataset \citep{abu2016youtube}, a large-scale collection of YouTube videos that also contain topical labels.
Since much of prior work in this area revolved around cooking videos, we aimed at sampling a significant proportion of our data from videos with cooking labels (specifically, ``Cooking'' and ``Recipe'').
Aside from the intentional bias regarding cooking videos, the rest of the videos were selected by randomly sampling non-cooking videos, including only those that were considered to be instructional videos by our human annotators.

Once candidate videos were identified, timeline annotations and descriptive tags were collected.
Our motivation was to enable downstream applications to allow navigating to specific content sections.  
Therefore, annotators were asked to identify the main steps in a video and mark their start time.  
They were also asked to produce a descriptive-yet-concise, free-text tag for each step (e.g., ``shaping the cookies'', ``removing any leftover glass'').
A subset of the videos has received more than one complete annotation (main steps plus tags).

The resulting \ldvm dataset consists of a total of 8,169 videos, of which 3,381 are cooking-related.
A total of 5,840 videos have received only one annotation, and have been designated as the training split.
Videos with more than one annotation have been designated as validation / test data.
Overall, there are 7.1 segments per video on average (max: 19).
Given the dataset design, descriptions are much shorter in length compared to \youcook: on average there are 2.97 words per tag (max: 16) --- 20\% of the captions are single-word, 22\% are two-words, and 25\% are three words.
Note that the average caption length is significantly shorter than for \youcook, which is not surprising given our motivation of providing short and concise timeline tags for video navigation.
We standardized the paraphrases among the top-20 most frequent captions.
For instance, \{``intro'', ``introduction''\}~$\to$~``intro''.
Otherwise, we have preserved the original tags as-is, even though additional paraphrasing most definitely exists.
Annotators were instructed to start and end the video with an opening and closing segment as possible.
As a result, the most frequent tag (post-standardization) in the dataset is ``intro'', which accounts for roughly 11\% of the 88,455 segments.  More details on the data collection process and additional analysis can be found in the Supplementary Material (Section \ref{sec:data_app}).

Overall, this results in 56,027 unique tags, with a vocabulary size of 12,509 token types over 88,455 segments.   
In this paper, we consider two variants: the full dataset (\ldvmmerged), and the cooking subset (\ldvmcooking).

\subsection{Pretraining Datasets: ASR+Video}

We consider two large-scale unannotated video datasets for pretraining, as described below. 
Time-stamped ASR tokens were obtained via YouTube Data API,\footnote{\url{https://developers.google.com/youtube/v3/docs/captions}}
and split into ASR segments if the timestamps of two consecutive words are more than 2 seconds apart, or if a segment is longer than a pre-specified max length (in our case, 320 words).
They were paired with concurrent video frames in the same segment.

\paragraph{\ytdatasub} 
We extract the cooking subset of \ytdata\citep{abu2016youtube} by taking, from its training split, videos with  ``Cooking'' or ``Recipe'' labels, and retain those with English ASR, subject to YouTube policies.
After preprocessing, we obtain 186K ASR+video segments with an average length of 64 words (24 seconds) per segment. 

\paragraph{\howto.}  This is based on the 1.2M YouTube instructional videos released by \newcite{miech2019howto100m}, covering a broad range of topics.  
After preprocessing, we obtain 7.99M ASR+video segments with an average of 78 words (28.7 seconds) per segment.

\subsection{Pretraining Datasets: CAP-style}

We also consider two text-only datasets for \textit{pretraining}, containing \textit{unpaired} instruction steps similar in style to the target captions.

\paragraph{\recipes} is a collection of 1M recipes scraped from a number of popular cooking websites~\citep{marin2019recipe1m+}. We use the sequence of instructions extracted for each recipe in this dataset, and treat each recipe step as a separate example during pretraining.
This results in 10,767,594 CAP-style segments, with 12.8 words per segment.

\paragraph{\wikihow} is a collection of 230,843 articles extracted from the WikiHow knowledge base ~\citep{koupaee2018wikihow}. Each article comes with a title starting with ``How to''.  Each associated step starts with a step summary (in bold) followed by a detailed explanation. 
We extract the all step summaries, resulting in 1,360,145 CAP-style segments, with 8.2 words per segment.
Again, each instruction step is considered as a separate example during pretraining.

\subsection{Differences between Pretraining and Finetuning Datasets}

First, note that \textit{video segments} are defined differently for pretraining and finetuning datasets, and may not match exactly. For ASR+Video pretraining datasets, which are unsupervised, the segments are divided following a simple heuristic (e.g., two consecutive words more than 2 seconds apart), whereas for finetuning ASR+Video$\rightarrow$CAP datasets, which are supervised, the segments are defined by human annotators to correspond to instruction steps. 
Otherwise, the ASR data are relatively similar between pretraining and finetuning datasets, as both come from instructional videos and are in the style of spoken language.

Second, compared to the target captions in finetuning datasets, the CAP-like pretraining datasets are similar in spirit --- they all represent summaries of \textit{steps}, but they may differ in length, style and granularity.
In particular, the CAP-like pretraining datasets are closer in style to captions in YouCook2, where annotators were instructed to produce a recipe-like description for each step.  This is reflected in their similar average length (YouCook2: 8.8 words, \recipes: 12.8 words, \wikihow: 8.2 words); whereas captions in \ldvm are significantly shorter (2.97 words on average).

Despite these differences --- some are inevitable due to the unsupervised nature of pretraining datasets --- the pretraining data is very helpful for our task as shown in the experimental results.

\section{Method}
\label{sec:model}
To model segment-level caption generation, we adopt MASS-style pretraining~\cite{song2019mass} with Transformer~\cite{vaswani2017attention} as the backbone architecture. For both pre-training and fine-tuning objectives, we have considered two variants: text-only 
and multi-modal.
They are summarized in Table~\ref{table:objectives} and more details are given below.

\subsection{Separate-Modality Architecture \label{sec:arch}}

\begin{figure*}
    \centering
    \includegraphics[width=\linewidth]{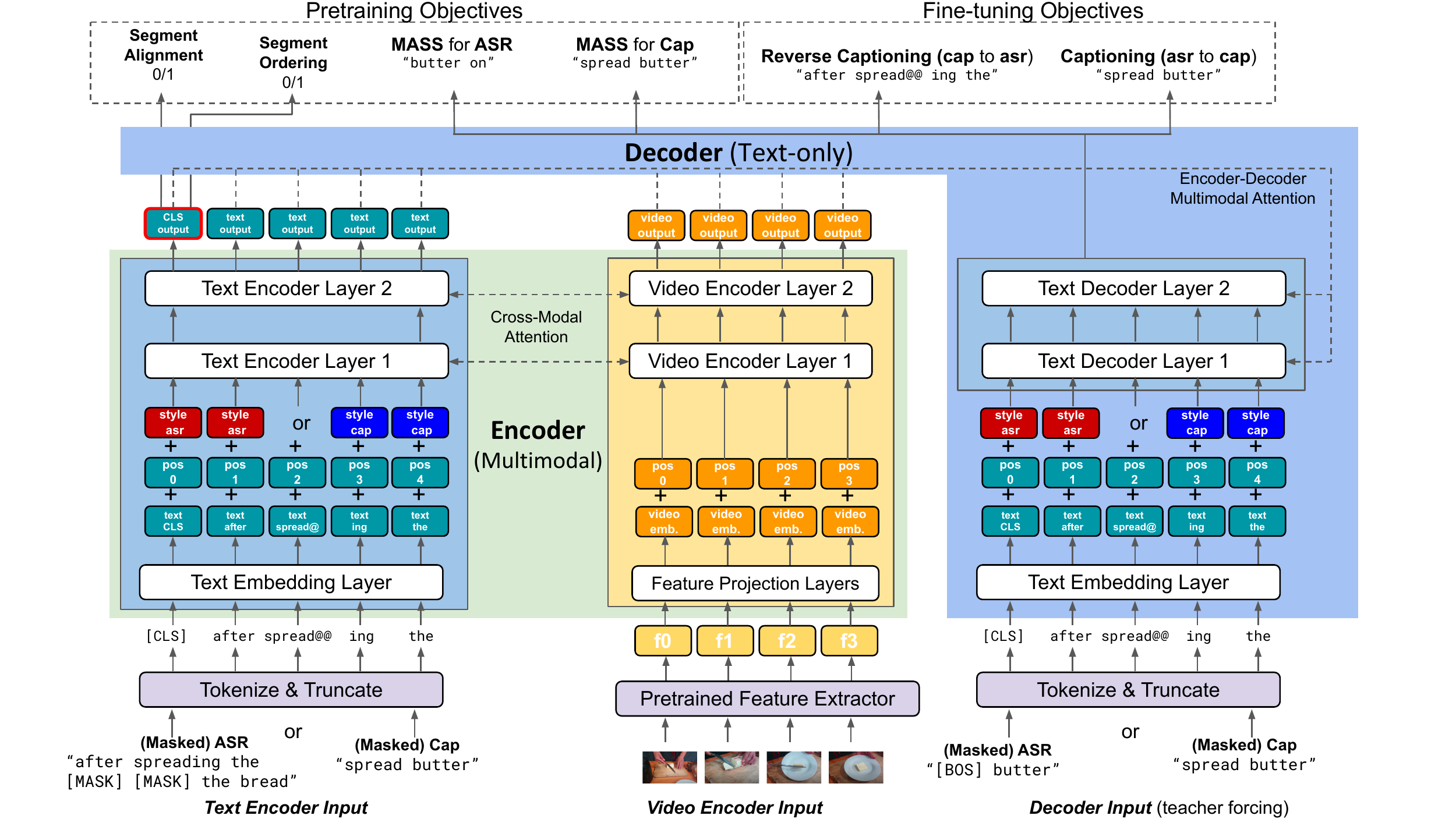}
    \caption{A diagram for the separate-modality architecture. It consists of a two-stream (text and video inputs) encoder with cross-modal attention and a text-only decoder, jointly trained using the MASS objective.
    }
    \label{fig:architecture}
\end{figure*}

Both ASR tokens and video segment features are given as input in the multimodal variants. We consider an architecture with a separate transformer for each modality (text or video), see Figure \ref{fig:architecture} for details. 
When available, the text and video encoders attend to each other at every layer using cross-modal attention, as in ViLBERT~\cite{lu2019vilbert}.
The text decoder attends over the final-layer output of both encoders.
We discuss in more detail the differences between using a separate-modality architecture vs. a vanilla-Transformer approach for all modalities in Appendix~\ref{sec:separatedvsconcatenated}.

The inputs to the text encoder is the sum of three components: 
text token embeddings, positional embeddings and the corresponding style embeddings,\footnote{This is similar to the way language-ID embeddings are used in machine translation.} depending on the style of the text (ASR or Caption-like).
The inputs to the video encoder could be either precomputed frame-level 2D CNN features or 3D CNN features, pretrained on the Kinetics~\cite{Carreira2017QuoVA, Kay2017TheKH} data set.
The visual features are projected with fully-connected layers to the same dimension as the text embeddings.

The main architecture we consider is a 2-layer encoder (E2), 6-layer decoder (D6) Transformer.
We use \textbf{E2D6} to refer to the text-only version, and \textbf{E2vidD6} to refer to the multimodal version with an active video encoder.
We also experiment with E2D2 and E2vidD2 (2-layer decoder).\footnote{We found in a preliminary study that using 6-layer encoders did not improve performance for our application.}

\subsection{Pretraining with Text-only MASS}

Text-only pretraining is essentially the unsupervised learning of the style transfer between ASR-style and caption-style texts using {\em unpaired} data sources: ASR strings from video segments in \youtube or \howto; and CAP-style instruction steps
found in \recipes or \howto.
Just like using MASS for unsupervised machine translation involves pretraining the model on unpaired monolingual datasets,
we alternate between \ASRtoASR and \CAPtoCAP MASS steps during our pretraining stage,
which does not require the ``source'' (ASR) and ``target'' (CAP-style) data to be aligned.

In an \ASRtoASR step, we mask a random subsequence of the ASR and feed the masked ASR to the text encoder. The text decoder must reconstruct the hidden subsequence while attending to the encoder output. A \CAPtoCAP step works similarly by trying to reconstruct a masked sequence of a CAP-style text.  The encoder and decoder are trained jointly using teacher-forcing on the decoder. We denote this text-only strategy as \textbf{MASS} in the experiments.

\subsection{Pretraining with Multimodal MASS}

During multimodal pretraining, 
we alternate between text-only \CAPtoCAP MASS steps and multimodal MASS steps.
During each multimodal MASS step \textsc{asr}+video$\to$\textsc{asr}, we feed a masked ASR to the text-encoder and the co-occurring video features to the video-encoder.
The text decoder must reconstruct the masked ASR subsequence.
We denote this pretraining strategy as \textbf{MASSvid} in the experiments.
This trains cross-modal attention between the text-encoder and video-encoder at every layer, 
jointly with the text decoder that attends to the output layer of both the text and video encoders.\footnote{
In preliminary experiments, we had attempted to directly adapt the MASS objective~\citep{song2019mass} to video reconstruction --- by masking a subsequence of the input video and making the video decoder reconstruct the input using the Noise Constrastive Estimator Loss~\citep{sun2019contrastive}. 
Due to limited success, we did not further pursue this approach.
%
}

To force more cross-modal attention between encoder and decoder,
we also investigate a strategy of hiding the text-encoder output from the decoder for some fraction of training examples.
We refer to this strategy as \textbf{MASSdrop} in the experiments.

\subsection{Pretraining with Alignment and Ordering~Tasks}

We also explore encoder-only multimodal pretraining strategies.
We take the last-layer representation for the CLS (beginning of sentence) token from the encoder, and add a multi-layer perceptron on top of it for binary predictions (Figure ~\ref{fig:architecture}).
Given a pair of ASR and video segment, we train the encoder to predict the following objectives:
\vspace{-0.5em}
\begin{itemize}
\setlength\itemsep{-0em}
  \item Segment-Level \textbf{Alignment}. An (ASR, video) pair is \textit{aligned} if they occur in the same pretraining segment; negative examples are constructed by sampling pairs from the same video but at least 2 segments away.
  \item Segment-Level \textbf{Ordering}.  We sample (ASR, video) pairs that are at least 2 segments away, and train the model to predict whether the ASR occurs before or after the video clip.
\end{itemize}



During this \textbf{MASSalign} pretraining stage, we alternate between two text-only MASS steps (\CAPtoCAP, \ASRtoASR) and the two binary predictions (\textbf{Alignment} and \textbf{Ordering}) described above.

\subsection{Finetuning on Video Captioning}


For text-only finetuning, we feed ASR to the text encoder and the decoder has to predict the corresponding CAP (\textsc{asr}$\to$\textsc{cap}). For multimodal finetuning, we also feed additional video representations to the video encoder (\textsc{asr}+video$\to$\textsc{cap}).  When finetuning a multimodal model from text-only pretraining, everything related to video (weights in the video encoder and any cross-modal attention modules) will be initialized randomly.
In addition to these \textit{uni-directional} (\textbf{\unimt}) finetuning, 
we also experiment with several variants of \textit{bidirectional}  (\textbf{\bimt}) finetuning (Table \ref{table:objectives}). For instance, adding \textsc{cap}$\to$\textsc{asr} (predicting ASR from CAP) to text-only finetuning.
In the experiments, we find some variants of bidirectional finetuning beneficial whether training from scratch or finetuning from a pretrained model.

\begin{table}
\centering
\begin{adjustbox}{width=\linewidth}
\setlength\tabcolsep{4pt} 
\begin{tabular}{lllll}
\toprule
\multicolumn{4}{c}{\textit{Pretraining Objectives}}\\
\midrule
\textbf{Name}  & \textbf{T} & \textbf{V} & {\textbf{Input}$\to$\textbf{Output}} \\
\midrule
MASS            & \cmark & \xmark & CAP$\to$CAP, ASR$\to$ASR \\
MASSvid         & \cmark & \cmark & CAP$\to$CAP, ASR+video$\to$ASR \\
MASSdrop    & \cmark & \cmark &     CAP$\to$CAP, ASR+video$\to$ASR \\
\multirow{2}*{MASSalign}       & \multirow{2}*{\cmark} & \multirow{2}*{\cmark} &  CAP$\to$CAP, ASR$\to$ASR, \\ 
                                &&&     ASR+video$\to$$\lbrace 0,1\rbrace$\\
\midrule
\multicolumn{4}{c}{\textit{Finetuning Objectives}}\\
\midrule
\textbf{Name}  & \textbf{T} & \textbf{V} & {\textbf{Input}$\to$\textbf{Output}} \\
\midrule
\unimt & \cmark & \xmark &  ASR$\to$CAP\\
\bimt & \cmark & \xmark & ASR$\to$CAP, CAP$\to$ASR\\
\unimt & \cmark & \cmark & ASR+video$\to$CAP\\
\bimt & \cmark & \cmark &  ASR+video$\to$CAP, CAP$\to$ASR\\
\bimtalt & \cmark & \cmark &  ASR+video$\to$CAP, CAP+video$\to$ASR\\
\bottomrule
\end{tabular}
\end{adjustbox}
\caption{Pretraining and Fine-tuning objectives. For each strategy, \cmark~ indicates whether the text ({\bf T}) and video ({\bf V}) encoders are active, followed by a summary of training objectives involved in one training step.\label{table:objectives}}
\end{table}

\section{Experiments}
\begin{table*}[t]
\adjustbox{max width=\textwidth}{%
\begin{tabular}{lccrrrr}
\toprule
Method & Input & Pretraining & \bleu-4 & \meteor & \rouge & \cider\\
\midrule
Constant Pred~\citep{hessel2019case} & - & - & 2.70 & 10.30 & 21.70 & 0.15 \\
MART~\cite{Lei2020MARTMR} & Video & - &  8.00 & 15.90 & - & 0.36 \\
EMT~\cite{Zhou2018EndtoEndDV} & Video &  - & 4.38 & 11.55 & 27.44 & 0.38 \\
CBT~\cite{sun2019contrastive} & Video & Kinetics + \howto  & 5.12 & 12.97 & 30.44 & 0.64 \\
AT~\citep{hessel2019case} & ASR & - & 8.55 & 16.93 & 35.54 & 1.06\\
AT+Video~\citep{hessel2019case} & Video + ASR & - & 9.01 & 17.77 & 36.65 & 1.12\\
UniViLM \#1~\citep{luo2020univilm} & Video & - & 6.06 & 12.47 &31.48 & 0.64 \\
UniViLM \#2~\citep{luo2020univilm} & Video + ASR & - & 8.67 & 15.38 & 35.02 & 1.00 \\
UniViLM \#5~\citep{luo2020univilm} & Video + ASR & \howto & 10.42 & 16.93 & 38.02 & 1.20 \\
\midrule
\multicolumn{7}{l}{\textit{\O~Pretraining}}\\
E2D6-\bimt & ASR & - & 7.90 & 15.70 & 34.86 & 0.93\\
E2vidD6-\bimt & Video + ASR & - & 8.01 & 16.19 & 34.66 & 0.91\\

\midrule
\multicolumn{7}{l}{\textit{Text Pretraining}}\\
E2D6-MASS-\bimt & ASR & \youtube + \recipes & 10.60 & 17.42 & 38.08 & 1.20\\
E2vidD6-MASS-\bimt & Video + ASR & \youtube + \recipes & 11.47 & 17.70 & 38.80 & \bf{1.25} \\
\midrule
\multicolumn{7}{l}{\textit{Multimodal Pretraining}}\\
E2vidD6-MASSalign-\bimt & Video + ASR & \youtube + \recipes & 11.53 & 17.62 & \textbf{39.03} & 1.22\\
E2vidD6-MASSvid-\bimt & Video + ASR & \youtube + \recipes & \bf{12.04} & \bf{18.32} & \bf{39.03} & 1.23\\
E2vidD6-MASSdrop-\bimt & Video + ASR & \youtube + \recipes & 10.45 & 17.74 & 38.82 & 1.22\\
\midrule
Human~\citep{hessel2019case} & Video + ASR & - & 15.20 &25.90 &45.10 &3.80 \\
\bottomrule
\end{tabular}}
\caption{Segment-level captioning results on \youcook. 
We use \youtube and \recipes for pretraining.
The numbers for the related work (first group) are directly reported from the corresponding papers.
The last line is an estimate of human performance as reported by~\citet{hessel2019case}, and can be taken as a rough upper bound of the best performance achievable.
 \label{table:youcook2_short}}
\end{table*}

\newcommand{\colwidth}{0.075\textwidth}
\newcommand{\colalign}{>{\raggedleft}p{\colwidth}}
\begin{table*}
\centering

\adjustbox{max width=\linewidth}{%
\begin{tabular}{p{0.26\textwidth}p{0.15\textwidth}rrrrp{0.02\textwidth}rrrr}
\toprule
\multirow{2}* {Method} & \multirow{2}* {Input} & \multicolumn{4}{c}{\ldvmmerged}     &&   \multicolumn{4}{c}{\ldvmcooking} \\ \cline{3-6} \cline{8-11}
 &  & \footnotesize{\bleu-1} & \footnotesize{\meteor} & \footnotesize{\rouge} & \footnotesize{CIDEr} &&\footnotesize{\bleu-1} & \footnotesize{\meteor} & \footnotesize{\rouge} & \footnotesize{CIDEr} \\
\midrule
%
Constant baseline (``\textit{intro}'') & - & 1.42 & 3.32 & 11.15 & 0.28 && 1.16 & 2.93 & 10.21 & 0.25\\
\midrule
\multicolumn{7}{l}{ \textit{\O~Pretraining}}\\
E2D6-\bimt & ASR & 19.60 & 9.12 & 27.88 & 0.68 && 20.77 & 10.08 & 28.63 & 0.72\\
E2vidD6-\bimt & Video + ASR & 19.49 & 9.23 & 28.53 & 0.69 &&20.45 & 9.88 & 28.88 & 0.69\\
\midrule
\multicolumn{7}{l}{\textit{Text Pretraining}}\\
E2D6-MASS-\bimt & ASR & 21.93 & 10.60 & 30.45 & 0.79 && 24.79 & 12.25 & 32.40 & 0.88\\
E2vidD6-MASS-\bimt & Video + ASR & 22.44 & 10.83 & 31.27 & 0.81 && 24.22 & 12.22 & 32.60 & 0.89\\

\midrule

\multicolumn{7}{l}{\textit{Multimodal Pretraining}}\\
E2vidD6-MASSalign-\bimt & Video + ASR & 22.31 & 10.66 & 31.13 & 0.79  && \bf{24.92} & 12.25 & 33.09 & \bf{0.90} \\
E2vidD6-MASSvid-\bimt & Video + ASR  & \bf{22.45} & 10.76 & \textbf{31.49} & 0.80 && 24.87 & \bf{12.43} & 32.97 & \bf{0.90} \\
E2vidD6-MASSdrop-\bimt & Video + ASR & 22.37 & \bf{11.00} & 31.40 & \bf{0.82}  && 24.48 & 12.22 & \bf{33.10} & 0.89 \\
\midrule
Human & Video + ASR & 43.34 &33.56 &41.88 &1.26 && 41.61 & 32.50 & 41.59 & 1.21 \\
\bottomrule
\end{tabular}}
\caption{Segment-level captioning results on \ldvm. For \ldvmmerged we pretrain on \howto and \wikihow; for \ldvmcooking we pretrain on \youtube and \recipes. We report baseline scores for predicting the most common caption ``intro''. \label{table:ldvmmerged_short} We also estimate the human performance as a rough upper bound (details in Supplementary Material \ref{sec:data_app}; Table \ref{tab:human_performance}).}

\end{table*}

\subsection{Implementation Details }

We tokenize ASR and CAP inputs using byte-pair--encoding subwords~\citep{sennrich2015neural}, and truncate them to 240 subwords. We truncate video sequences to 40 frames (40 seconds of video), compute the 128-dim features proposed by~\citet{wang2014learning} (which we will refer to as \starburst features), and project them to the embedding space using a two-layer perceptron with layer normalization and GeLU activations. 

We instantiate the E2xDx models defined in Section~\ref{sec:arch} with 128-dimensional embeddings and 8 heads respectively for self-attention, encoder-decoder, and cross-modal attention modules. We define each epoch to be 3,125 iterations, where each iteration contains one repetition of each training step as represented in Table~\ref{table:objectives}. We pretrain for 200 epochs and finetune for 30 epochs.

For evaluation, we consider \bleu-4~\citep{papineni2002bleu}, \meteor~\citep{denkowski2014meteor}, \rouge~\citep{lin2004automatic} and CIDEr~\citep{vedantam2015cider} metrics.

Please refer to Appendix~\ref{app:implementation} for full implementation details, hyperparameters and computation cost.

\paragraph{Notes on \ldvm evaluation:} With the exception of \rouge, all other metrics are sensitive to short groundtruth.  67\% of the groundtruth tags in \ldvm have less than 4 words, where a perfect prediction will not yield a full score in, say, \bleu-4.  Thus, we focus mainly on \rouge, report \mbox{\bleu-1} instead of \bleu-4 for \ldvm, and provide the other two metrics only as reference points.

We had originally decided to use videos with multiple annotations as validation and test data, so that we could explore evaluation with multiple reference groundtruth captions.  But as annotators do not always yield the same set of segment boundaries, this became tricky.  Instead, we simply treat each segment as a separate instance with one single reference caption.  Note that all segments annotated for the same video will be in either validation or test to ensure no content overlap.


\subsection{Main Results}

We run several variants of our method on \youcook, \ldvmmerged and \ldvmcooking, using different architectures, modalities, pretraining datasets, pretraining and finetuning strategies.

\paragraph{Comparing with other methods on YouCook2} 
For YouCook2, we report our method alongside several methods from the literature~\citep{hessel2019case,sun2019videobert,Zhou2018EndtoEndDV,Lei2020MARTMR}, as well as state-of-the-art concurrent work~\citep{luo2020univilm}.
The related work is provided for reference and to give a ballpark estimate of the relative performance of each method, but results are not always strictly and directly comparable.
Beyond the usual sources of discrepancy in data processing, tokenization, or even different splits, an additional source of complication comes from the fact that videos are regularly deleted by content creators, causing video datasets to shrink over time.
Additionally, when comparing to other work incorporating pretraining, we could differ in (videos available in) pretraining datasets, segmentation strategies, etc.
To this end, we perform an extensive ablation study, which at least helps us to understand the effectiveness of different components in our approach.

\paragraph{Effect of pretraining}
The main experimental results for the three datasets we consider are summarized in Table \ref{table:youcook2_short} (\youcook) and Table \ref{table:ldvmmerged_short} (\ldvmmerged and \ldvmcooking).
Across all three datasets, the best performance is achieved by finetuning a multimodal captioning model under the {\em Multimodal Pretraining} condition.
For instance, on \youcook, E2vidD6-MASSvid-\bimt improves over the no-pretraining model E2vidD6-BiD by 4.37 \rouge, a larger improvement than UniViLM with pretraining (\#5) vs without (\#2) \cite{luo2020univilm}. 
This improvement also holds in \ldvmcooking (+4.22 in \rouge) and \ldvmmerged (+2.97 in \rouge).  
We do not observe consistent and significant trends among the different multimodal pretraining strategies: MASS pretraining with video (\textbf{MASSvid}), with video and droptext (\textbf{MASSdrop}), or with alignment tasks (\textbf{MASSalign}).\footnote{Limited improvement with MASSalign suggests that such alignment tasks are better suited for retrieval~\citep{luo2020univilm}.}
%
%
Furthermore, we observe that most of the pretraining improvement is achievable via text-only MASS pretraining.
Across all three datasets, while {\em Multimodal Pretraining} (E2vidD6-MASSvid-\bimt) is consistently better than {\em Text Pretraining} (E2vidD6-MASS-\bimt), the differences are quite small (under one \rouge point).

It's worthy noting that for MASSalign, the best validation accuracies for the pretraining tasks are reasonably high: for \ytdatasub, we observed 90\% accuracy for the alignment task, and 80\% for the ordering task (for \howto: 87\% and 71.4\%, respectively), where random guess would yield 50\%.  This suggests that our video features are reasonably strong, and the findings above are not due to weak visual representations.

\begin{table}[]
\adjustbox{max width=\linewidth}{%
\setlength\tabcolsep{3pt} 
\begin{tabular}{lrrrr}
\toprule
Method & \bleu-4 & \meteor & \rouge & CIDEr\\
\midrule
D2-\unimt & 10.84 & 17.39 & 38.24 & 1.16\\
D6-\unimt & 11.39 & 18.00 & 38.71 & 1.22\\

D2-\bimt & 11.38 & \bf{18.04} & 38.67 & 1.19\\
D6-\bimt & 11.47 & 17.70 & \bf{38.80} & \bf{1.25} \\

D6-\bimtalt & 11.07 & 17.68 & 38.43 & 1.22\\

D6-\bimt (S3D)  & \textbf{11.64} & 18.04 & 38.75 & 1.24\\

\bottomrule
\end{tabular}}
\caption{Ablation study on YouCook2. We finetune a multimodal captioning model (E2vid) with either 2-layer decoder (D2) or 6-layer decoder (D6) using \youtube/\recipes for MASS pretraining, combined with either unidirectional (\unimt) or bidirectional (\bimt) finetuning. We find no significant difference between using 2D and 3D features (marked as S3D).\label{table:ablation}}

\end{table}

\paragraph{Effect of other modeling choices}
We experiment with 2-layer decoder (D2) vs 6-layer decoder (D6), combined with either unidirectional fine-tuning (\textbf{\unimt}) or bidirectional fine-tuning (\textbf{\bimt}).
Table \ref{table:ablation} shows ablation results of the four possible combinations when finetuning a multimodal model using text-only pretraining on \youcook (a more complete list of results can be found in Appendix~\ref{sec:fullresults}, showing similar trends).
The D6x\bimt combination tends to yield the best performance, with the differences among the four configurations being relatively small (under one \rouge point).  
For visual features, we also explored using 3D features~\citep{xie2018rethinking} instead of 2D features during finetuning (with no pretraining or text-only pretraining), and do not find much difference in model performance on \youcook.
As a result, we use the simpler 2D features in our multimodal pretraining.  We leave more extensive experiments with visual features  as future work. 

\paragraph{Generalization implications}
An important motivation for constructing the \ldvm dataset and evaluating our models on it has been related to generalization.
Since the \youcook benchmark is restricted to a small number of cooking recipes, 
there is little to be understood about how well models trained and evaluated on it generalize.
In contrast, the \ldvm benchmark has a much wider coverage (for both cooking-related videos and general instructional videos), and no imposed topic overlap between train/dev/test.
As such, there are two findings here that are relevant with respect to generalization: (a) the absolute performance of the models on the \ldvm benchmark is quite high (ROUGE-L scores above 0.30 are usually indicative of decent performance), and (b) the performance on \ldvm vs. \youcook is clearly lower (31.5 ROUGE-L vs. 39.0 ROUGE-L, reflecting the increased difficulty of the new benchmark), but it is maximized under similar pretraining and finetuning conditions, which allows us to claim that the resulting models generalize well and are quite robust over a wide variety of instructional videos.

\section{Conclusions}

Motivated to improve information-seeking capabilities for videos, we have collected and annotated a new dense video captioning dataset, \ldvm, which is larger with higher diversity
compared to \youcook.
We investigated several multimodal pretraining strategies for segment-level video captioning, and conducted extensive ablation studies.
We concluded that MASS-style pretraining is the most decisive factor in improving the performance on all the benchmarks used.
Even more to the point, our results indicate that most of the performance can be attributed to leveraging the ASR signal. 
We achieve new state-of-the-art results on the \youcook benchmark, and establish strong performance baselines for the new \ldvm benchmark, which can be used as starting points for driving more progress in this direction.

\section*{Acknowledgements}

We send warm thanks to Ashish Thapliyal for helping the first author debug his code and navigate the computing infrastructure, and to Sebastian Goodman for his technical help (and lightning fast responses!). 
We also thank the anonymous reviewers for their comments and suggestions.

\bibliography{mmpt.bib}

\begin{thebibliography}{53}
\expandafter\ifx\csname natexlab\endcsname\relax\def\natexlab#1{#1}\fi

\bibitem[{Abu-El-Haija et~al.(2016)Abu-El-Haija, Kothari, Lee, Natsev,
  Toderici, Varadarajan, and Vijayanarasimhan}]{abu2016youtube}
Sami Abu-El-Haija, Nisarg Kothari, Joonseok Lee, Paul Natsev, George Toderici,
  Balakrishnan Varadarajan, and Sudheendra Vijayanarasimhan. 2016.
\newblock Youtube-8m: A large-scale video classification benchmark.
\newblock \emph{arXiv preprint arXiv:1609.08675}.

\bibitem[{Alayrac et~al.(2016)Alayrac, Bojanowski, Agrawal, Josef~Sivic, and
  Lacoste-Julien}]{Alayrac2016cvpr}
Jean-Baptiste Alayrac, Piotr Bojanowski, Nishant Agrawal, Ivan~Laptev
  Josef~Sivic, and Simon Lacoste-Julien. 2016.
\newblock Unsupervised learning from narrated instruction videos.
\newblock In \emph{2016 IEEE Conference on Computer Vision and Pattern
  Recognition (CVPR)}. IEEE.

\bibitem[{Carreira and Zisserman(2017)}]{Carreira2017QuoVA}
Jo{\~a}o Carreira and Andrew Zisserman. 2017.
\newblock Quo vadis, action recognition? a new model and the kinetics dataset.
\newblock \emph{2017 IEEE Conference on Computer Vision and Pattern Recognition
  (CVPR)}, pages 4724--4733.

\bibitem[{Chen et~al.(2019)Chen, Li, Yu, Kholy, Ahmed, Gan, Cheng, and
  Liu}]{chen2019uniter}
Yen-Chun Chen, Linjie Li, Licheng Yu, Ahmed~El Kholy, Faisal Ahmed, Zhe Gan,
  Yu~Cheng, and Jingjing Liu. 2019.
\newblock Uniter: Learning universal image-text representations.
\newblock \emph{arXiv preprint arXiv:1909.11740}.

\bibitem[{Damen et~al.(2018)Damen, Doughty, Farinella, Fidler, Furnari,
  Kazakos, Moltisanti, Munro, Perrett, Price, and Wray}]{DamenECCV2018}
Dima Damen, Hazel Doughty, Giovanni~Maria Farinella, Sanja Fidler, Antonino
  Furnari, Evangelos Kazakos, Davide Moltisanti, Jonathan Munro, Toby Perrett,
  Will Price, and Michael Wray. 2018.
\newblock Scaling egocentric vision: The {EPIC-KITCHENS} dataset.
\newblock In \emph{European Conference on Computer Vision (ECCV)}.

\bibitem[{Denkowski and Lavie(2014)}]{denkowski2014meteor}
Michael Denkowski and Alon Lavie. 2014.
\newblock Meteor universal: Language specific translation evaluation for any
  target language.
\newblock In \emph{Proceedings of the ninth workshop on statistical machine
  translation}, pages 376--380.

\bibitem[{Devlin et~al.(2018)Devlin, Chang, Lee, and
  Toutanova}]{devlin2018bert}
Jacob Devlin, Ming-Wei Chang, Kenton Lee, and Kristina Toutanova. 2018.
\newblock Bert: Pre-training of deep bidirectional transformers for language
  understanding.
\newblock \emph{arXiv preprint arXiv:1810.04805}.

\bibitem[{Dong et~al.(2019)Dong, Yang, Wang, Wei, Liu, Wang, Gao, Zhou, and
  Hon}]{UniLM}
Li~Dong, Nan Yang, Wenhui Wang, Furu Wei, Xiaodong Liu, Yu~Wang, Jianfeng Gao,
  Ming Zhou, and Hsiao{-}Wuen Hon. 2019.
\newblock Unified language model pre-training for natural language
  understanding and generation.
\newblock \emph{arXiv preprint arXiv:1905.03197}.

\bibitem[{Hessel et~al.(2019)Hessel, Pang, Zhu, and Soricut}]{hessel2019case}
Jack Hessel, Bo~Pang, Zhenhai Zhu, and Radu Soricut. 2019.
\newblock A case study on combining asr and visual features for generating
  instructional video captions.
\newblock In \emph{Proceedings of CoNLL}.

\bibitem[{Kay et~al.(2017)Kay, Carreira, Simonyan, Zhang, Hillier,
  Vijayanarasimhan, Viola, Green, Back, Natsev, Suleyman, and
  Zisserman}]{Kay2017TheKH}
Will Kay, Jo{\~a}o Carreira, Karen Simonyan, Brian Zhang, Chloe Hillier,
  Sudheendra Vijayanarasimhan, Fabio Viola, Tim Green, Trevor Back, Apostol
  Natsev, Mustafa Suleyman, and Andrew Zisserman. 2017.
\newblock The kinetics human action video dataset.
\newblock \emph{ArXiv}, abs/1705.06950.

\bibitem[{Kim et~al.(2014)Kim, Nguyen, Weir, Guo, Miller, and
  Gajos}]{kim2014crowdsourcing}
Juho Kim, Phu~Tran Nguyen, Sarah Weir, Philip~J Guo, Robert~C Miller, and
  Krzysztof~Z Gajos. 2014.
\newblock Crowdsourcing step-by-step information extraction to enhance existing
  how-to videos.
\newblock In \emph{CHI}.

\bibitem[{Koupaee and Wang(2018)}]{koupaee2018wikihow}
Mahnaz Koupaee and William~Yang Wang. 2018.
\newblock Wikihow: A large scale text summarization dataset.
\newblock \emph{arXiv preprint arXiv:1810.09305}.

\bibitem[{Krishna et~al.(2017)Krishna, Hata, Ren, Fei-Fei, and
  Niebles}]{krishna2017densecaptioning}
Ranjay Krishna, Kenji Hata, Frederic Ren, Li~Fei-Fei, and Juan~Carlos Niebles.
  2017.
\newblock Dense-captioning events in videos.
\newblock In \emph{International Conference on Computer Vision (ICCV)}.

\bibitem[{Lan et~al.(2020)Lan, Chen, Goodman, Gimpel, Sharma, and
  Soricut}]{Lan2020ALBERTAL}
Zhenzhong Lan, Mingda Chen, Sebastian Goodman, Kevin Gimpel, Piyush Sharma, and
  Radu Soricut. 2020.
\newblock Albert: A lite bert for self-supervised learning of language
  representations.
\newblock \emph{ArXiv}, abs/1909.11942.

\bibitem[{Lei et~al.(2020)Lei, Wang, Shen, Yu, Berg, and
  Bansal}]{Lei2020MARTMR}
Jie Lei, Liwei Wang, Yelong Shen, Dong Yu, Tamara~L. Berg, and Mohit Bansal.
  2020.
\newblock Mart: Memory-augmented recurrent transformer for coherent video
  paragraph captioning.
\newblock In \emph{The 58th Annual Meeting of the Association for Computational
  Linguistics}.

\bibitem[{Lewis et~al.(2019)Lewis, Liu, Goyal, Ghazvininejad, Mohamed, Levy,
  Stoyanov, and Zettlemoyer}]{Lewis2019BARTDS}
Mike Lewis, Yinhan Liu, Naman Goyal, Marjan Ghazvininejad, Abdelrahman Mohamed,
  Omer Levy, Ves Stoyanov, and Luke Zettlemoyer. 2019.
\newblock Bart: Denoising sequence-to-sequence pre-training for natural
  language generation, translation, and comprehension.
\newblock \emph{ArXiv}, abs/1910.13461.

\bibitem[{Li et~al.(2019)Li, Duan, Fang, Jiang, and Zhou}]{li2019unicoder}
Gen Li, Nan Duan, Yuejian Fang, Daxin Jiang, and Ming Zhou. 2019.
\newblock Unicoder-vl: A universal encoder for vision and language by
  cross-modal pre-training.
\newblock \emph{arXiv preprint arXiv:1908.06066}.

\bibitem[{Lin and Och(2004)}]{lin2004automatic}
Chin-Yew Lin and Franz~Josef Och. 2004.
\newblock Automatic evaluation of machine translation quality using longest
  common subsequence and skip-bigram statistics.
\newblock In \emph{Proceedings of the 42nd Annual Meeting on Association for
  Computational Linguistics}, page 605. Association for Computational
  Linguistics.

\bibitem[{Liu et~al.(2019)Liu, Ott, Goyal, Du, Joshi, Chen, Levy, Lewis,
  Zettlemoyer, and Stoyanov}]{Roberta}
Yinhan Liu, Myle Ott, Naman Goyal, Jingfei Du, Mandar Joshi, Danqi Chen, Omer
  Levy, Mike Lewis, Luke Zettlemoyer, and Veselin Stoyanov. 2019.
\newblock Roberta: {A} robustly optimized {BERT} pretraining approach.
\newblock \emph{arXiv preprint arxiv:1907.11692}.

\bibitem[{Lu et~al.(2019)Lu, Batra, Parikh, and Lee}]{lu2019vilbert}
Jiasen Lu, Dhruv Batra, Devi Parikh, and Stefan Lee. 2019.
\newblock Vilbert: Pretraining task-agnostic visiolinguistic representations
  for vision-and-language tasks.
\newblock In \emph{Advances in Neural Information Processing Systems}, pages
  13--23.

\bibitem[{Luo et~al.(2020)Luo, Ji, Shi, Huang, Duan, Li, Chen, and
  Zhou}]{luo2020univilm}
Huaishao Luo, Lei Ji, Botian Shi, Haoyang Huang, Nan Duan, Tianrui Li, Xilin
  Chen, and Ming Zhou. 2020.
\newblock Univilm: A unified video and language pre-training model for
  multimodal understanding and generation.
\newblock \emph{arXiv preprint arXiv:2002.06353}.

\bibitem[{Margulieux et~al.(2012)Margulieux, Guzdial, and
  Catrambone}]{margulieux2012subgoal}
Lauren~E Margulieux, Mark Guzdial, and Richard Catrambone. 2012.
\newblock Subgoal-labeled instructional material improves performance and
  transfer in learning to develop mobile applications.
\newblock In \emph{Conference on International Computing Education Research}.

\bibitem[{Marin et~al.(2019)Marin, Biswas, Ofli, Hynes, Salvador, Aytar, Weber,
  and Torralba}]{marin2019recipe1m+}
Javier Marin, Aritro Biswas, Ferda Ofli, Nicholas Hynes, Amaia Salvador, Yusuf
  Aytar, Ingmar Weber, and Antonio Torralba. 2019.
\newblock Recipe1m+: A dataset for learning cross-modal embeddings for cooking
  recipes and food images.
\newblock \emph{IEEE transactions on pattern analysis and machine
  intelligence}.

\bibitem[{Miech et~al.(2020)Miech, Alayrac, Smaira, Laptev, and
  Josef~Sivic}]{Miech2020cvpr}
Antoine Miech, Jean-Baptiste Alayrac, Lucas Smaira, Ivan Laptev, and
  Andrew~Zisserman Josef~Sivic. 2020.
\newblock End-to-end learning of visual representations from uncurated
  instructional videos.
\newblock In \emph{IEEE Conference on Computer Vision and Pattern Recognition
  (CVPR)}. IEEE.

\bibitem[{Miech et~al.(2019)Miech, Zhukov, Alayrac, Tapaswi, Laptev, and
  Sivic}]{miech2019howto100m}
Antoine Miech, Dimitri Zhukov, Jean-Baptiste Alayrac, Makarand Tapaswi, Ivan
  Laptev, and Josef Sivic. 2019.
\newblock Howto100m: Learning a text-video embedding by watching hundred
  million narrated video clips.
\newblock In \emph{Proceedings of the IEEE International Conference on Computer
  Vision}, pages 2630--2640.

\bibitem[{O'Neil-Hart(2018)}]{oneilhart2018}
Celie O'Neil-Hart. 2018.
\newblock Why you should lean into how-to content in 2018.
\newblock
  \url{www.thinkwithgoogle.com/advertising-channels/video/self-directed-learning-youtube/}.
\newblock Accessed: 2019-09-03.

\bibitem[{Pan et~al.(2020)Pan, Cai, Huang, Lee, Gaidon, Adeli, and
  Niebles}]{pan2020cvpr}
Boxiao Pan, Haoye Cai, De-An Huang, Kuan-Hui Lee, Adrien Gaidon, Ehsan Adeli,
  and Juan~Carlos Niebles. 2020.
\newblock Spatio-temporal graph for video captioning with knowledge
  distillation.
\newblock In \emph{IEEE Conference on Computer Vision and Pattern Recognition
  (CVPR)}. IEEE.

\bibitem[{Papineni et~al.(2002)Papineni, Roukos, Ward, and
  Zhu}]{papineni2002bleu}
Kishore Papineni, Salim Roukos, Todd Ward, and Wei-Jing Zhu. 2002.
\newblock Bleu: a method for automatic evaluation of machine translation.
\newblock In \emph{Proceedings of the 40th annual meeting on association for
  computational linguistics}, pages 311--318. Association for Computational
  Linguistics.

\bibitem[{Radford et~al.(2018)Radford, Narasimhan, Salimans, and
  Sutskever}]{GPT2018}
Alec Radford, Karthik Narasimhan, Tim Salimans, and Ilya Sutskever. 2018.
\newblock Improving language understanding by generative pre-training.
\newblock
  \emph{https://s3-us-west-2.amazonaws.com/openai-assets/research-covers/language-unsupervised/language\_understanding\_paper.pdf}.

\bibitem[{Raffel et~al.(2019)Raffel, Shazeer, Roberts, Lee, Narang, Matena,
  Zhou, Li, and Liu}]{Raffel2019ExploringTL}
Colin Raffel, Noam Shazeer, Adam Roberts, Katherine Lee, Sharan Narang, Michael
  Matena, Yanqi Zhou, Wei Li, and Peter~J. Liu. 2019.
\newblock Exploring the limits of transfer learning with a unified text-to-text
  transformer.
\newblock \emph{ArXiv}, abs/1910.10683.

\bibitem[{Sanabria et~al.(2018)Sanabria, Caglayan, Palaskar, Elliott, Barrault,
  Specia, and Metze}]{how2-dataset}
Ramon Sanabria, Ozan Caglayan, Shruti Palaskar, Desmond Elliott, Lo{\"{\i}}c
  Barrault, Lucia Specia, and Florian Metze. 2018.
\newblock How2: {A} large-scale dataset for multimodal language understanding.
\newblock \emph{arXiv preprint arXiv:1811.00347}.

\bibitem[{Sennrich et~al.(2015)Sennrich, Haddow, and
  Birch}]{sennrich2015neural}
Rico Sennrich, Barry Haddow, and Alexandra Birch. 2015.
\newblock Neural machine translation of rare words with subword units.
\newblock \emph{arXiv preprint arXiv:1508.07909}.

\bibitem[{Shi et~al.(2019)Shi, Ji, Liang, Duan, Chen, Niu, and
  Zhou}]{shi2019dense}
Botian Shi, Lei Ji, Yaobo Liang, Nan Duan, Peng Chen, Zhendong Niu, and Ming
  Zhou. 2019.
\newblock Dense procedure captioning in narrated instructional videos.
\newblock In \emph{Proceedings of the 57th Annual Meeting of the Association
  for Computational Linguistics}, pages 6382--6391.

\bibitem[{Song et~al.(2019)Song, Tan, Qin, Lu, and Liu}]{song2019mass}
Kaitao Song, Xu~Tan, Tao Qin, Jianfeng Lu, and Tie-Yan Liu. 2019.
\newblock Mass: Masked sequence to sequence pre-training for language
  generation.
\newblock \emph{arXiv preprint arXiv:1905.02450}.

\bibitem[{Su et~al.(2019)Su, Zhu, Cao, Li, Lu, Wei, and Dai}]{su2019vl}
Weijie Su, Xizhou Zhu, Yue Cao, Bin Li, Lewei Lu, Furu Wei, and Jifeng Dai.
  2019.
\newblock Vl-bert: Pre-training of generic visual-linguistic representations.
\newblock \emph{arXiv preprint arXiv:1908.08530}.

\bibitem[{Sun et~al.(2019{\natexlab{a}})Sun, Baradel, Murphy, and
  Schmid}]{sun2019contrastive}
Chen Sun, Fabien Baradel, Kevin Murphy, and Cordelia Schmid.
  2019{\natexlab{a}}.
\newblock Contrastive bidirectional transformer for temporal representation
  learning.
\newblock \emph{arXiv preprint arXiv:1906.05743}.

\bibitem[{Sun et~al.(2019{\natexlab{b}})Sun, Myers, Vondrick, Murphy, and
  Schmid}]{sun2019videobert}
Chen Sun, Austin Myers, Carl Vondrick, Kevin Murphy, and Cordelia Schmid.
  2019{\natexlab{b}}.
\newblock Videobert: A joint model for video and language representation
  learning.
\newblock In \emph{Proceedings of the IEEE International Conference on Computer
  Vision}, pages 7464--7473.

\bibitem[{Tan and Bansal(2019)}]{tan2019lxmert}
Hao Tan and Mohit Bansal. 2019.
\newblock Lxmert: Learning cross-modality encoder representations from
  transformers.
\newblock \emph{arXiv preprint arXiv:1908.07490}.

\bibitem[{Vaswani et~al.(2017{\natexlab{a}})Vaswani, Shazeer, Parmar,
  Uszkoreit, Jones, Gomez, Kaiser, and Polosukhin}]{Vaswani2017AttentionIA}
Ashish Vaswani, Noam Shazeer, Niki Parmar, Jakob Uszkoreit, Llion Jones,
  Aidan~N. Gomez, Lukasz Kaiser, and Illia Polosukhin. 2017{\natexlab{a}}.
\newblock Attention is all you need.
\newblock \emph{ArXiv}, abs/1706.03762.

\bibitem[{Vaswani et~al.(2017{\natexlab{b}})Vaswani, Shazeer, Parmar,
  Uszkoreit, Jones, Gomez, Kaiser, and Polosukhin}]{vaswani2017attention}
Ashish Vaswani, Noam Shazeer, Niki Parmar, Jakob Uszkoreit, Llion Jones,
  Aidan~N Gomez, {\L}ukasz Kaiser, and Illia Polosukhin. 2017{\natexlab{b}}.
\newblock Attention is all you need.
\newblock In \emph{Advances in neural information processing systems}, pages
  5998--6008.

\bibitem[{Vedantam et~al.(2015)Vedantam, Lawrence~Zitnick, and
  Parikh}]{vedantam2015cider}
Ramakrishna Vedantam, C~Lawrence~Zitnick, and Devi Parikh. 2015.
\newblock Cider: Consensus-based image description evaluation.
\newblock In \emph{Proceedings of the IEEE conference on computer vision and
  pattern recognition}, pages 4566--4575.

\bibitem[{Wang et~al.(2014)Wang, Song, Leung, Rosenberg, Wang, Philbin, Chen,
  and Wu}]{wang2014learning}
Jiang Wang, Yang Song, Thomas Leung, Chuck Rosenberg, Jingbin Wang, James
  Philbin, Bo~Chen, and Ying Wu. 2014.
\newblock Learning fine-grained image similarity with deep ranking.
\newblock In \emph{Proceedings of the IEEE Conference on Computer Vision and
  Pattern Recognition}, pages 1386--1393.

\bibitem[{Wang et~al.(2019)Wang, Wu, Chen, Li, fang Wang, and
  Wang}]{Wang2019VaTeXAL}
Xin Wang, Jiawei Wu, Junkun Chen, Lei Li, Yuan fang Wang, and William~Yang
  Wang. 2019.
\newblock Vatex: A large-scale, high-quality multilingual dataset for
  video-and-language research.
\newblock \emph{2019 IEEE/CVF International Conference on Computer Vision
  (ICCV)}, pages 4580--4590.

\bibitem[{Weir et~al.(2015)Weir, Kim, Gajos, and
  Miller}]{weir2015learnersourcing}
Sarah Weir, Juho Kim, Krzysztof~Z Gajos, and Robert~C Miller. 2015.
\newblock Learnersourcing subgoal labels for how-to videos.
\newblock In \emph{CSCW}.

\bibitem[{Xie et~al.(2018)Xie, Sun, Huang, Tu, and Murphy}]{xie2018rethinking}
Saining Xie, Chen Sun, Jonathan Huang, Zhuowen Tu, and Kevin Murphy. 2018.
\newblock Rethinking spatiotemporal feature learning: Speed-accuracy trade-offs
  in video classification.
\newblock In \emph{Proceedings of the European Conference on Computer Vision
  (ECCV)}, pages 305--321.

\bibitem[{YouTubeBlog(2017)}]{youtubeblog}
YouTubeBlog. 2017.
\newblock You know what’s cool? a billion hours.
\newblock
  \url{https://youtube.googleblog.com/2017/02/you-know-whats-cool-billion-hours.html}.
\newblock Accessed: 2020-06-23.

\bibitem[{ZenithMedia(2019)}]{zenithmedia}
ZenithMedia. 2019.
\newblock Online video viewing to reach 100 minutes a day in 2021.
\newblock
  \url{https://www.zenithmedia.com/online-video-viewing-to-reach-100-minutes-a-day-in-2021/}.
\newblock Accessed: 2020-06-23.

\bibitem[{Zhang et~al.(2020)Zhang, Shi, Yuan, Li, Wang, Hu, and
  Zha}]{zhang2020cvpr}
Ziqi Zhang, Yaya Shi, Chunfeng Yuan, Bing Li, Peijin Wang, Weiming Hu, and
  Zhengjun Zha. 2020.
\newblock Object relational graph with teacher-recommended learning for video
  captioning.
\newblock In \emph{IEEE Conference on Computer Vision and Pattern Recognition
  (CVPR)}. IEEE.

\bibitem[{Zheng et~al.(2020)Zheng, Wang, and Tao}]{zheng2020cvpr}
Qi~Zheng, Chaoyue Wang, and Dacheng Tao. 2020.
\newblock Syntax-aware action targeting for video captioning.
\newblock In \emph{IEEE Conference on Computer Vision and Pattern Recognition
  (CVPR)}. IEEE.

\bibitem[{Zhou et~al.(2018{\natexlab{a}})Zhou, Xu, and Corso}]{zhou2018towards}
Luowei Zhou, Chenliang Xu, and Jason~J Corso. 2018{\natexlab{a}}.
\newblock Towards automatic learning of procedures from web instructional
  videos.
\newblock In \emph{Thirty-Second AAAI Conference on Artificial Intelligence}.

\bibitem[{Zhou et~al.(2018{\natexlab{b}})Zhou, Xu, and Corso}]{youcook2dataset}
Luowei Zhou, Chenliang Xu, and Jason~J. Corso. 2018{\natexlab{b}}.
\newblock You{CookII} dataset.
\newblock
  \url{http://youcook2.eecs.umich.edu/static/YouCookII/youcookii\_readme.pdf}.
\newblock Accessed: 2020-06-23.

\bibitem[{Zhou et~al.(2018{\natexlab{c}})Zhou, Zhou, Corso, Socher, and
  Xiong}]{Zhou2018EndtoEndDV}
Luowei Zhou, Yingbo Zhou, Jason~J. Corso, Richard Socher, and Caiming Xiong.
  2018{\natexlab{c}}.
\newblock End-to-end dense video captioning with masked transformer.
\newblock \emph{2018 IEEE/CVF Conference on Computer Vision and Pattern
  Recognition}, pages 8739--8748.

\bibitem[{Zhu and Yang(2020)}]{actbert2020cvpr}
Linchao Zhu and Yi~Yang. 2020.
\newblock Actbert: Learning global-local video-text representations.
\newblock In \emph{IEEE Conference on Computer Vision and Pattern Recognition
  (CVPR)}. IEEE.

\end{thebibliography}
\bibliographystyle{acl_natbib}

\clearpage

\appendix

\section{Appendix}
Supplementary Material for ``Multimodal Pretraining for Dense Video Captioning''.
\subsection{The \ldvm dataset}
\label{sec:data_app}
\paragraph{Sampling video for annotation.}
The goal of the ViTT dataset design is to mirror topic distribution in the ``wild''. Therefore, instead of starting from specific how-to instructions and searching for corresponding videos, we sampled videos from the validation set of the \ytdata dataset \citep{abu2016youtube},  
a large-scale collection of YouTube videos with topical labels, subject to YouTube policies.

Exclusion criteria were lack of English ASR and the topic label ``Game''. The latter was motivated by the fact that in this type of videos, the visual information predominantly features video games, while the ViTT dataset was intended to contain only  videos with real-world human actions.
Cooking videos can be easily identified by sampling videos that came with ``Cooking'' or ``Recipe'' topic labels.  
Given the convenience and the fact that much of prior work in this area had focused on cooking videos, approximately half of the dataset was designed to include cooking videos only, while the remaining videos would be randomly sampled non-cooking videos, as long as they were verified as instructional by human annotators.  



\paragraph{Annotation process}
Annotators were presented with a video alongside its timestamped, automatic transcription shown in sentence-length paragraphs.
They were asked to watch the video and first judge whether the video was instructional. For the purpose of our dataset, we determine that a video is instructional if it focuses on real-world human actions that are accompanied by procedural language explaining what is happening on screen, in reasonable details. Also for our purposes, instructional videos need to be grounded in real life, with a real person in the video exemplifying the action being verbally described.
  
For videos judged to be instructional, annotators were then asked to:
\begin{itemize}
\item Delimit the main segments of the video.
\item Determine their start time if different from the automatically suggested start time (explained below).
\item Provide a label summarizing or explaining the segment. 
\end{itemize}

\paragraph{Annotation guidelines}
Annotators were instructed to identify video segments with two potential purposes:
\begin{itemize}
\item Allow viewers to jump straight to the start of a segment for rewatch.
\item Present viewers with an index to decide whether to watch the video in full or directly skip to the segment of interest.
\end{itemize}

Our guidelines suggested a range of five to ten segments as long as the the structure and content of the video permitted. For short videos, the direction was to prioritize quality over quantity and to only define those segments that formed the narrative structure of the video, even if the resulting number of segments was below 5. 

To help annotators determine segment start times, transcriptions were shown in ``sentences'' --- we expected that sentence start times might be good candidates for segment start times.   We obtained sentence boundaries automatically as follows.
Given the stream of timestamped ASR tokens for a video, we first separated them into blocks by breaking two consecutive tokens whenever they were more than 2 seconds apart.  We then used a punctuation prediction model to identify sentence boundaries in each resulting block.  Each sentence was shown with the timestamp corresponding to its first token.  Annotators were advised that transcriptions had been automatically divided into paragraphs that may or may not correspond to a video segment --- if they decided that a segment started from a particular sentence, they could choose to use the start time of the sentence as the start time for the segment, or, if needed, they could put in an adjusted start time instead.

Once the start time had been identified, annotators were asked to provide a free-text label to summarize each segment.   We instructed the annotators to use nouns or present participles (-ing form of verbs) to write the labels for the video segments, whenever possible.  Additionally, we asked that the labels be succinct while descriptive, using as few words as possible to convey as much information as possible.

\paragraph{Data statistics and post-processing}
The resulting dataset consists of 8,169 instructional videos that received segment-level annotations, of which 3,381 are cooking-related.
Overall there are an average of 7.1 segments per video (max: 19). Given our instructions, the descriptions are much shorter in lengths compared to a typical captioning dataset:  on average there are 2.97 words per description (max: 16);  20\% of the captions are single-word, 22\% are two-words, and 25\% are three words.  We refer to these descriptions as ``tags'' given how short they are.
 
When possible, annotators were also asked to start and end the video with an opening and closing segment.
As a result, most annotations start with an introduction segment: this accounts for roughly 11\% of the 88455 segments in the dataset (``intro'': 8\%, ``introduction'': 2.3\%).  Note that while ``intro'' and ``introduction'' are clearly paraphrases of each other, an automatic metric will penalize a model predicting ``intro'' when the groundtruth is ``introduction''.  
Similarly, the ending segment was described in several varieties: ``outro'': 3.4\%, ``closing'': 1\%, ``closure'', ``conclusion'', ``ending'', ```end of video'': each under 1\%.
Penalizing paraphrases of the ground truth is an inherent weakness of automatic metrics.  To mitigate this, we decided to reduce the chance of this happening for the most frequent tags in the dataset.  That is, in our experiments, we identified three groups of tags among the top-20 most frequent tags, and standardized them as follows.
\begin{table}[h!]
    \centering
    \begin{tabular}{|l|l|} \hline 
intro     &  intro, introduction, opening \\ \hline
outro     &  outro, closing, closure, conclusion, \\
          &  ending, end of video, video closing \\ \hline 
result    &  finished result, final result, results \\ \hline
\end{tabular}
    \caption{Standardization of top tags}
    \label{tab:tag_normalization}
\end{table}

Note that this does not mean we can solve this problem as a classification task like in visual question answering (VQA):
overall, there are 56,027 unique tags with a vocabulary size of 12,509 for the 88,455 segments; 51,474 tags appeared only once in the dataset, making it infeasible to reduce the segment-level captioning problem into a pure classification task.  Table \ref{tab:top10_tags} shows the top 10 most frequent tags after standardization.

\begin{table}[]
    \centering
    \begin{tabular}{l|c} \hline
Tag & \% of segments \\ \hline
intro & 11.4 \\
outro & 6.6 \\
result & 0.9 \\
ingredients & 0.8 \\
listing ingredients & 0.2 \\
supplies & 0.2 \\
mixing ingredients & 0.2 \\
materials & 0.1 \\
what you'll need & 0.1 \\
lining the eyes & 0.1 \\    \hline
    \end{tabular}
    \caption{10 most frequent tags after standardization.}
    \label{tab:top10_tags}
\end{table}

\paragraph{Estimate of human performance.}
A subset of the candidate videos were given to three annotators\footnote{A small set were unintentionally given to six annotators.}, to help us understand variations in human annotations.
5,840 videos received dense captioning from exactly one annotator and were used as training data.
Videos with more than one annotation were used as validation / test data.  Note that not all the videos with multiple timeline annotations have exactly three sets of them --- in fact, 1368 videos received 3-way segment-level annotations.  This is because not all annotators agreed on whether a video was instructional.   Computing annotator agreement for the annotated timelines is non-trivial.  Here we focus on an estimate of tagging agreement when a pair of annotators agreed over the segment start time.  Specifically, we go through each video that received multiple segment-level annotations.  For each segment where two annotators chose the same ASR sentence as its starting point, we take the tags they produced for this segment and consider one of them as groundtruth, the other as prediction, and add that into our pool of (groundtruth, prediction) pairs.  We can then compute standard automatic evaluations metrics over this pool.  The results are as follows.
\begin{table}[h!]
    \centering
    \begin{tabular}{ccccc} \hline
        BLEU-1 & METEOR & ROUGE-L & CIDEr   \\ \hline
        43.34 &33.56 &41.88 &1.26    \\ \hline
    \end{tabular}
    \caption{Estimate of human performance for the segment-level captioning on \ldvmmerged (computed over 7528 pairs).}
    \label{tab:human_performance}
\end{table}
\begin{table}[h!]
    \centering
    \begin{tabular}{ccccc} \hline
        BLEU-1 & METEOR & ROUGE-L & CIDEr   \\ \hline
        41.61 & 32.50 & 41.59 & 1.21    \\ \hline
    \end{tabular}
    \caption{Estimate of human performance for the segment-level captioning on \ldvmcooking  (computed over 2511 pairs).}
    \label{tab:human_performance}
\end{table}

Note that METEOR, and CIDEr scores are both penalized by the lack of n-grams for higher n.  That is, when both groundtruth and prediction are single-word, say, ``intro'', this pair will not receive a full score from any of these metrics.  But the \rouge score is in the same ballpark as estimate of human performance in prior work \cite{hessel2019case}.  One might note that perhaps this pool of label pairs contains a higher share of ``intro'', since annotators might be more likely to agree over where an opening segment starts.  Indeed, 20\% of the time, one of the tags is ``intro''.  Interestingly, in spite of  standardization of top tags, 14\% of the time one tag is ``intro'', the other tag is {\em not} ``intro'': they can be less frequent paraphrases (e.g., ``welcoming'', ``greeting'', ``opening and welcoming'') or something semantically different (e.g., ``using dremel tool'').

\subsection{Separated vs. Concatenated-Modality Architecture
\label{sec:separatedvsconcatenated}}

Prior work has explored both concatenating different modalities and feeding them into the same multimodal Transformer encoder \citep{sun2019videobert, hessel2019case}, as well as separating them into unimodal transformers \citep{sun2019contrastive,lu2019vilbert}.
We opt for the separated architecture because it offers more flexibility.
First, the concatenated architecture requires embedding the text and video features into the same space.
When the video features are projected using a simple network, there is no guarantee that we can meaningfully project them into the text embedding space.  VideoBERT~\citep{sun2019videobert} gives more flexibility to the video embeddings by  quantizing video features and learning an embedding for each codeword. However, the quantization step has subsequently been claimed to be detrimental~\citep{sun2019contrastive}.
Moreover, the concatenated architecture uses the same sets of forward and attention weights to process text and video, and performs layer normalization jointly between the two modalities, which is not necessarily meaningful.
Finally, the separated architecture makes it easy to switch between variable length text-only, video-only, or text+video modalities, whereas concatenated architectures might rely on separating tokens, modalities embeddings, and using fixed sequence lengths~\citep{luo2020univilm}.


\subsection{Additional Implementation Details~\label{app:implementation}}

We optimize all models on a nVidia v100 GPU using the Adam optimizer with inverse square root schedule, batch size 32, warm-up period of 4,000 iterations, and maximum learning rate of $0.0001$, following MASS~\citep{song2019mass}. The positional embeddings are initialized randomly. We use dropout and attention dropout with probabilities $0.1$.
 With E2vidD6, pretraining takes 3-6 days depending on the objective and bidirectional finetuning takes up to 1.5 days, however those times could be improved by optimizing the data pipeline.

\subsection{Example Predictions}

We show examples of \textcolor{green!70!black}{\textbf{good}} and \textcolor{red}{\textbf{bad}} predictions on \youcook (Figure~\ref{fig:supplementary_youcook} and \ldvmmerged (Figure~\ref{fig:supplementary_ldvm_good} and~\ref{fig:supplementary_ldvm_bad}).
The captions are generated by E2vidD6-BiD (no pretraining) and E2vidD6-MASS-BiD (text-only MASS pretraining).

\begin{figure*}
\centering
{
\small
\begin{adjustbox}{max width=\linewidth}
\begin{tabular}{p{0.19\textwidth} p{0.19\textwidth} p{0.19\textwidth} p{0.19\textwidth} p{0.19\textwidth} }
\toprule
\textbf{Sample Frame} & \textbf{Ground Truth} & \textbf{\O-Pretraining} & \textbf{MASS-Pretraining} & Comments \\
\midrule
\ex{plots/youcook2_segments/line_000792}%
{crush and grate the garlic}%
{grate garlic and add to bowl \good}%
{crush ginger and garlic\good } & ginger is correct despite not appearing in ground truth.\\

\ex{plots/youcook2_segments/line_000038}%
{crimp shut with fork}%
{place the filling on the wrapper \ok}%
{seal the edges of the wrapper \good} & pretrained model is more specific \\

\ex{plots/youcook2_segments/line_000002}%
{place wings on the baking sheet and cook flipping}%
{bake the pizza in the oven \bad}%
{cook the wings on the grill \good} & only pretrained model predicted correct food\\

\ex{plots/youcook2_segments/line_000010}%
{add the pork back into the hot oil  }%
{add the rice to the pot \bad}%
{place the meat on the pan \good}  & \O~model hallucinates the rice and pot\\
\ex{plots/youcook2_segments/line_000017}%
{add thyme bay leaves onion and clam juice and boil the mixture}%
{add diced tomatoes tomato puree and mix well \bad }%
{add thyme thyme onion and clam juice to the pot and stir \ok}  & \O~hallucinates a lot of nonexistent ingredients\\

\ex{plots/youcook2_segments/line_000770}%
{cook bacon in a pot with oil and pepper}%
{add chopped tomatoes to pan and stir \bad}%
{add bacon and stir \ok } & both models missed oil and pepper (not mentioned in ASR)\\

\ex{plots/youcook2_segments/line_000053}%
{pour dressing on top of the salad and toss}%
{add dressing to the bowl \good}%
{serve the soup over the salad \bad }  & pretrained model referred to dressing as ``soup''\\

\ex{plots/youcook2_segments/line_000791}%
{slice the ginger into pieces}%
{slice a celery \bad}%
{slice the chicken \bad } & both models had wrong ingredients (ASR segment does not mention what is being sliced) \\

\bottomrule
\end{tabular}
\end{adjustbox}
}
\caption{Example good and bad predictions on \youcook. The pretrained model is generally but not always better. Note that there are no ``intro'' or ``outro''-like labels on \youcook because the dataset was specifically curated to only contain actual recipe steps.}
\label{fig:supplementary_youcook}
\end{figure*}

\begin{figure*}
\centering
{
\small
\begin{adjustbox}{max width=\linewidth}
\begin{tabular}{p{0.19\textwidth} p{0.19\textwidth} p{0.19\textwidth} p{0.19\textwidth} p{0.19\textwidth} }
\toprule
\textbf{Sample Frame} & \textbf{Ground Truth} & \textbf{\O-Pretraining} & \textbf{MASS-Pretraining} & Comments \\
\midrule

\ex{plots/ldvm_segments/line_000184}%
{tightening extra loop}%
{tightening the loop \good}%
{tightening the loop \good} & both models perform well\\

\ex{plots/ldvm_segments/line_010080}%
{adding eyeshadow}%
{blending eye shadow \good}%
{applying eye shadow \good} & both models perform well\\

\ex{plots/ldvm_segments/line_000135}%
{showcasing the finished look}%
{showing finished look\good}%
{showing finished look\good} & both models perform well\\

\ex{plots/ldvm_segments/line_000484}%
{rolling and folding the clay}%
{rolling and blending \ok}%
{rolling and folding the clay \good} & MASS is a bit more specific\\

\ex{plots/ldvm_segments/line_001212}%
{highlighting brow bone}%
{applying eye shadow \ok}%
{brushing on the brows\good} & MASS is a bit more specific\\

\ex{plots/ldvm_segments/line_010053}%
{covering the chicken and cooking}%
{cooking the bread \bad}%
{cooking the chicken \good} & only MASS got the right ingredient\\

\ex{plots/ldvm_segments/line_000091}%
{connecting spray hose and sprayer}%
{connecting the new cover \ok}%
{connecting the valve \good} & spray hose is more specific than valve\\

\ex{plots/ldvm_segments/line_000113}%
{implementing second layer}%
{showing finished product \ok}%
{showing second layer \good} & MASS is more specific\\

\ex{plots/ldvm_segments/line_000114}%
{making decorative trim}%
{cutting the edges \good}%
{cutting the fabric \good} & both models yield good predictions\\

\ex{plots/ldvm_segments/line_000201}%
{checking bleach container}%
{outro \bad}%
{checking the container \good} & MASS is a bit more specific\\

\ex{plots/ldvm_segments/line_000174}%
{demonstrating the flip}%
{checking the battery \bad}%
{flipping the board \good} & \O~model got influenced by car mechanics tutorials\\

\ex{plots/ldvm_segments/line_010023}%
{tilting board}%
{setting up the oven \bad} %
{turning the board \good} & \O~overfitted on cooking videos\\

\bottomrule
\end{tabular}
\end{adjustbox}
}
\caption{Example \textcolor{green}{\textbf{good}} predictions on \ldvmmerged (Part 1). The pretrained model is generally but not always better. \label{fig:supplementary_ldvm_good}}
\label{fig:supplementary_youcook}
\end{figure*}

\begin{figure*}
\centering
{
\small
\begin{adjustbox}{max width=\linewidth}
\begin{tabular}{p{0.19\textwidth} p{0.19\textwidth} p{0.19\textwidth} p{0.19\textwidth} p{0.19\textwidth} }
\toprule
\textbf{Sample Frame} & \textbf{Ground Truth} & \textbf{\O-Pretraining} & \textbf{MASS-Pretraining} & Comments \\
\midrule

\ex{plots/ldvm_segments/line_010102}%
{securing the bar in place}%
{removing the cover \bad}%
{checking for the other side \bad} & predictions are not specific enough\\

\ex{plots/ldvm_segments/line_010105}%
{starting with unlocking bars}%
{opening the box \bad}%
{pulling the car on \bad} & predictions are incorrect or not specific enough\\

\ex{plots/ldvm_segments/line_000083}%
{demonstrating technique}%
{attaching paper \bad}%
{stamping paper \good} & the technique is about stamping the paper\\

\ex{plots/ldvm_segments/line_000037}%
{spritzing in additional water}%
{pouring water into the water \ok}%
{adding water to water \ok} & understandable but ungrammarly\\

\ex{plots/ldvm_segments/line_000094}%
{checking for leaks}%
{checking for the new new new new new new new new new new new new new new new \bad}%
{checking the process \ok} & \O~got into a loop, MASS not specific enough\\

\ex{plots/ldvm_segments/line_010030}%
{displaying materials needed}%
{intro \bad}%
{removing paste \ok} & prediction makes sense because narrator is displaying thermal paste remover\\

\ex{plots/ldvm_segments/line_000023}%
{sketching on the swirls}%
{drawing the lines \good}%
{drawing on the eyes \bad} & pretrained model overfitted on makeup tutorials\\

\ex{plots/ldvm_segments/line_000286}%
{crimping wire and completing project}%
{attaching the screws \bad}%
{attaching the wire to the wire \ok} & both models have trouble with the concept of crimping a wire\\

\ex{plots/ldvm_segments/line_000236}%
{cutting with guide line}%
{cutting the top of the top of the top of the top of the top of the top \bad}%
{explaining process \ok} & \O~model got into a loop, MASS model is not specific enough\\

\bottomrule
\end{tabular}
\end{adjustbox}
}
\caption{Example \textbf{\textcolor{orange}{ok}} and \textbf{\textcolor{red}{bad}} predictions on \ldvm (Part 2). The pretrained model is generally but not always better. \label{fig:supplementary_ldvm_bad}}
\label{fig:supplementary_youcook}
\end{figure*}

\subsection{Full result tables}
\label{sec:fullresults}
We present here tables with all the ablation results that we run.
There are two main takeaway messages from the results involving the pretraining approach:
(a) the accuracy improvements, as measured across all the metrics we use, indicate the value of using a pretraining approach to this problem, specifically one that is capable of leveraging the ASR signals at both pretraining and finetuning stages, and
(b) the training speedup achieved from pretraining is impressive, as a pretrained model converges much faster than training from scratch. This is especially visible on \ldvmmerged where finetuning after MASS pretraining reaches best \rouge score at epoch 2, whereas it takes around 11 epochs to converge when training from scratch.

\begin{table*}
\adjustbox{max width=\textwidth}{%
\begin{tabular}{lccrrrr}
\toprule
Method & Input & Pretraining & BLEU-4 & METEOR & ROUGE-L & CIDEr\\
\midrule
Constant Pred~\citep{hessel2019case} & - & - & 2.70 & 10.30 & 21.70 & 0.15 \\
MART~\cite{Lei2020MARTMR} & Video & - &  8.00 & 15.90 & - & 0.36 \\
DPC~\cite{shi2019dense} & Video + ASR & - & 2.76 & 18.08 & - & - \\
EMT~\cite{Zhou2018EndtoEndDV} & Video &  - & 4.38 & 11.55 & 27.44 & 0.38 \\
CBT~\cite{sun2019contrastive} & Video & Kinetics + \howto  & 5.12 & 12.97 & 30.44 & 0.64 \\
AT~\citep{hessel2019case} & ASR & - & 8.55 & 16.93 & 35.54 & 1.06\\
AT+Video~\citep{hessel2019case} & Video + ASR & - & 9.01 & 17.77 & 36.65 & 1.12\\
UniViLM \#1~\citep{luo2020univilm} & Video & - & 6.06 & 12.47 &31.48 & 0.64 \\
UniViLM \#2~\citep{luo2020univilm} & Video + ASR & - & 8.67 & 15.38 & 35.02 & 1.00 \\
UniViLM \#5~\citep{luo2020univilm} & Video + ASR & \howto & 10.42 & 16.93 & 38.02 & 1.20 \\
\midrule
\multicolumn{7}{l}{\textit{\O~Pretraining}}\\
E2D2-\unimt & ASR & - & 7.42 & 15.15 & 33.26 & 0.85\\
E2D6-\unimt & ASR & - & 7.88 & 15.29 & 34.10 & 0.87\\
E2D2-\bimt & ASR & - & 6.85 & 15.64 & 34.26 & 0.91\\
E2D6-\bimt & ASR & - & 7.90 & 15.70 & 34.86 & 0.93\\
E2vidD2-\unimt & Video + ASR & - & 7.47 & 15.11 & 34.77 & 0.90\\
E2vidD6-\unimt & Video + ASR & - & 7.61 & 15.57 & 34.28 & 0.89\\
E2vidD2-\bimt & Video + ASR & - & 8.39 & 15.36 & 34.54 & 0.91\\
E2vidD6-\bimt & Video + ASR & - & 8.01 & 16.19 & 34.66 & 0.91\\
E2vidD2-\bimtalt & Video + ASR & - & 8.12 & 15.83 & 34.83 & 0.93\\
E2vid,D6-\bimtalt & Video + ASR & - & 7.70 & 16.11 & 34.78 & 0.91\\

E2vidD2-\bimt (S3D) & Video + ASR & - & 8.04 & 16.17 & 36.01 & 0.96\\
E2vidD6-\bimt (S3D) & Video + ASR & - & 7.91 & 16.28 & 35.23 & 0.93\\

\midrule
\multicolumn{7}{l}{\textit{Text Pretraining}}\\
E2D2-MASS-\unimt & ASR & \youtube + \recipes & 10.52 & 17.14 & 37.39 & 1.14\\
E2D6-MASS-\unimt & ASR & \youtube + \recipes & 10.72 & 17.74 & 37.85 & 1.17\\

E2D2-MASS-\bimt & ASR & \youtube + \recipes & 10.84 & 17.44 & 37.20 & 1.13\\
E2D6-MASS-\bimt & ASR & \youtube + \recipes & 10.60 & 17.42 & 38.08 & 1.20\\
E2vidD2-MASS-\unimt & Video + ASR & \youtube + \recipes & 10.84 & 17.39 & 38.24 & 1.16\\
E2vidD6-MASS-\unimt & Video + ASR & \youtube + \recipes & 11.39 & 18.00 & 38.71 & 1.22\\

E2vidD2-MASS-\bimt & Video + ASR & \youtube + \recipes & 11.38 & 18.04 & 38.67 & 1.19\\
E2vidD6-MASS-\bimt & Video + ASR & \youtube + \recipes & 11.47 & 17.70 & 38.80 & \bf{1.25} \\

E2vid,D2-MASS-\bimtalt & Video + ASR & \youtube + \recipes & 11.49 & 17.85 & 38.60 & 1.18\\

E2vid,D6-MASS-\bimtalt & Video + ASR & \youtube + \recipes & 11.07 & 17.68 & 38.43 & 1.22\\

E2vidD2-MASS-\bimt (S3D) & Video + ASR & \youtube + \recipes & 11.13 & 17.71 & 38.57 & 1.12\\
E2vidD6-MASS-\bimt (S3D) & Video + ASR & \youtube + \recipes & 11.64 & 18.04 & 38.75 & 1.24\\

\midrule
\multicolumn{7}{l}{\textit{Multimodal Pretraining}}\\
E2vidD2-MASSalign-\bimt & Video + ASR & \youtube + \recipes & 11.54 & 17.57 & 37.70 & 1.15\\
E2vidD6-MASSalign-\bimt & Video + ASR & \youtube + \recipes & 11.53 & 17.62 & \textbf{39.03} & 1.22\\
E2vidD2-MASSvid-\bimt & Video + ASR & \youtube + \recipes & 11.17 & 17.71 & 38.32 & 1.17\\
E2vidD6-MASSvid-\bimt & Video + ASR & \youtube + \recipes & \bf{12.04} & \bf{18.32} & \bf{39.03} & 1.23\\
E2vidD2-MASSdrop-\bimt & Video + ASR & \youtube + \recipes & 11.21 & 17.99 & 38.72 & 1.23\\
E2vidD6-MASSdrop-\bimt & Video + ASR & \youtube + \recipes & 10.45 & 17.74 & 38.82 & 1.22\\
\midrule
Human~\citep{hessel2019case} & Video + ASR & - & 15.20 &25.90 &45.10 &3.80 \\
\bottomrule
\end{tabular}}
\caption{Video Captioning Results on YouCook2. We use \ytdatasub/\recipes for pretraining. All video features are \starburst~\citep{wang2014learning} except when marked as S3D~\citep{xie2018rethinking}. \label{table:youcook2}}
\end{table*}

\begin{table*}
\centering
\adjustbox{max width=\linewidth}{%
\begin{tabular}{lccrrrr}
\toprule
Method & Input & Pretraining & BLEU-1 & METEOR & ROUGE-L & CIDEr\\
\midrule

Constant baseline (``\textit{intro}'') & - & - & 1.42 & 3.32 & 11.15 & 0.28\\
\midrule
\multicolumn{7}{l}{ \textit{\O~Pretraining}}\\
E2D2-\unimt & ASR & - & 17.94 & 8.55 & 27.06 & 0.64\\
E2D6-\unimt & ASR & - & 18.91 & 8.96 & 27.80 & 0.67\\
E2D2-\bimt & ASR & - & 18.81 & 8.82 & 27.63 & 0.65\\
E2D6-\bimt & ASR & - & 19.60 & 9.12 & 27.88 & 0.68\\
E2vidD2-\unimt & Video + ASR & - & 18.94 & 8.99 & 28.05 & 0.67\\
E2vidD6-\unimt & Video + ASR & - & 19.29 & 9.15 & 27.97 & 0.69\\
E2vidD2-\bimt & Video + ASR & - & 19.37 & 9.21 & 28.56 & 0.69\\
E2vidD6-\bimt & Video + ASR & - & 19.49 & 9.23 & 28.53 & 0.69\\
\midrule
\multicolumn{7}{l}{\textit{Text Pretraining}}\\
E2D2-MASS-\unimt & ASR & \howto + \wikihow & 21.53 & 10.24 & 29.95 & 0.77\\
E2D6-MASS-\unimt & ASR & \howto + \wikihow & 22.09 & 10.58 & 30.67 & 0.79\\

E2D2-MASS-\bimt & ASR & \howto + \wikihow & 20.73 & 10.20 & 30.15 & 0.76\\
E2D6-MASS-\bimt & ASR & \howto + \wikihow & 21.93 & 10.60 & 30.45 & 0.79\\
E2vidD2-MASS-\unimt & Video + ASR & \howto + \wikihow & 21.46 & 10.45 & 30.56 & 0.78\\
E2vidD6-\unimt & Video + ASR & \howto + \wikihow & 22.21 & 10.75 & 30.86 & 0.81\\

E2vidD2-MASS-\bimt & Video + ASR & \howto + \wikihow & 21.78 & 10.64 & 30.72 & 0.79\\

E2vidD6-MASS-\bimt & Video + ASR & \howto + \wikihow & 22.44 & 10.83 & 31.27 & 0.81 \\

\midrule

\multicolumn{7}{l}{\textit{Multimodal Pretraining}}\\
E2vidD2-MASSalign-\bimt & Video + ASR & \howto + \wikihow & 22.07 & 10.33 & 30.60 & 0.77\\
E2vidD6-MASSalign-\bimt & Video + ASR & \howto + \wikihow & 22.31 & 10.66 & 31.13 & 0.79\\
E2vidD2-MASSvid-\bimt & Video + ASR & \howto + \wikihow & 22.15 & 10.75 & 31.06 & 0.80\\
E2vidD6-MASSvid-\bimt & Video + ASR & \howto + \wikihow & \bf{22.45} & 10.76 & \textbf{31.49} & 0.80\\
E2vidD2-MASSdrop-\bimt & Video + ASR & \howto + \wikihow & 21.84 & 10.55 & 31.10 & 0.79\\
E2vidD6-MASSdrop-\bimt & Video + ASR & \howto + \wikihow & 22.37 & \bf{11.00} & 31.40 & \bf{0.82}\\
\midrule
Human estimate & Video + ASR & - & 43.34 &33.56 &41.88 &1.26\\
\bottomrule
\end{tabular}}
\caption{Video captioning results on \ldvmmerged. We use \howto/\wikihow for pretraining. We also estimate human performance (details in Appendix~\ref{sec:data_app}; Table \ref{tab:human_performance}). \label{table:ldvmmerged}}

\end{table*}

\begin{table*}
\centering
\adjustbox{max width=\linewidth}{%
\begin{tabular}{lccrrrr}
\toprule
Method & Input & Pretraining & BLEU-1 & METEOR & ROUGE-L & CIDEr\\
\midrule
Constant baseline (``\textit{intro}'') & - & - & 1.16 & 2.93 & 10.21 & 0.25\\
\midrule
\multicolumn{7}{l}{\textit{\O~Pretraining}}\\
E2D2-\unimt & ASR & - & 19.73 & 9.43 & 27.95 & 0.69\\
E2D6-\unimt & ASR & - & 20.24 & 9.93 & 28.59 & 0.71\\
E2D2-\bimt & ASR & - & 19.73 & 9.72 & 27.92 & 0.68\\
E2D6-\bimt & ASR & - & 20.77 & 10.08 & 28.63 & 0.72\\
E2vidD2-\unimt & Video + ASR & - & 19.97 & 9.75 & 28.30 & 0.69\\
E2vidD6-\unimt & Video + ASR & - & 20.46 & 9.93 & 28.62 & 0.69\\
E2vidD2-\bimt & Video + ASR & - & 20.60 & 10.08 & 29.45 & 0.71\\
E2vidD6-\bimt & Video + ASR & - & 20.45 & 9.88 & 28.88 & 0.69\\
\midrule
\multicolumn{7}{l}{\textit{Text Pretraining}}\\
E2D2-MASS-\unimt & ASR & \youtube + \recipes & 22.89 & 11.53 & 31.62 & 0.84\\
E2D6-MASS-\unimt & ASR & \youtube + \recipes & 24.47 & 12.22 & 32.51 & 0.90\\

E2D2-MASS-\bimt & ASR & \youtube + \recipes & 22.75 & 11.63 & 31.54 & 0.84\\
E2D6-MASS-\bimt & ASR & \youtube + \recipes & 24.79 & 12.25 & 32.40 & 0.88\\
E2vidD2-MASS-\unimt & Video + ASR & \youtube + \recipes &  23.86 & 11.85 & 32.32 & 0.86\\
E2vidD6-MASS-\unimt & Video + ASR & \youtube + \recipes & 24.32 & 12.32 & 32.90 & 0.90\\

E2vidD2-MASS-\bimt & Video + ASR & \youtube + \recipes & 22.93 & 11.68 & 32.15 & 0.87\\
E2vidD6-MASS-\bimt & Video + ASR & \youtube + \recipes & 24.22 & 12.22 & 32.60 & 0.89\\
\midrule
\multicolumn{7}{l}{\textit{Multimodal Pretraining}}\\


E2vidD2-MASSalign-\bimt & Video + ASR & \youtube + \recipes & 24.02 & 11.91 & 32.73 & 0.86\\

E2vidD6-MASSalign-\bimt & Video + ASR & \youtube + \recipes & \bf{24.92} & 12.25 & 33.09 & \bf{0.90}\\
E2vidD2-MASSvid-\bimt & Video + ASR & \youtube + \recipes & 24.15 & 12.10 & 32.96 & 0.88\\
E2vidD6-MASSvid-\bimt & Video + ASR & \youtube + \recipes & 24.87 & \bf{12.43} & 32.97 & \bf{0.90} \\

E2vidD2-MASSdrop-\bimt & Video + ASR & \youtube + \recipes & 23.70 & 12.01 & 32.71 & 0.88\\
E2vidD6-MASSdrop-\bimt & Video + ASR & \youtube + \recipes & 24.48 & 12.22 & \bf{33.10} & 0.89\\
\midrule
Human estimate & Video + ASR & - & 41.61 & 32.50 & 41.59 & 1.21 \\
\bottomrule
\end{tabular}}
\caption{Video captioning results on \ldvmcooking. We use \youtube and \recipes for optional pretraining. \label{table:ldvmcooking}}

\end{table*}

\end{document}